\documentclass{article}
\usepackage{ijcai24}

\usepackage{times}
\usepackage{soul}
\usepackage{url}
\usepackage[hidelinks]{hyperref}
\usepackage[utf8]{inputenc}
\usepackage[small]{caption}
\usepackage{graphicx}
\usepackage{amsmath}
\usepackage{amsthm}
\usepackage{booktabs}
\usepackage{algorithm}
\usepackage{algorithmic} 
\usepackage[switch]{lineno}

\usepackage[toc,page]{appendix}

\newtheorem{theorem}{Theorem}

\title{Learning Logic Programs by Discovering Higher-Order Abstractions}

\author{
C\'{e}line Hocquette$^1$
\and
Sebastijan Duman\v{c}i\'{c}$^2$\and
Andrew Cropper$^{1}$
\affiliations
$^1$University of Oxford\\
$^2$TU Delft\\
\emails
\{celine.hocquette, andrew.cropper\}@cs.ox.ac.uk;
s.dumancic@tudelft.nl
}

\usepackage[utf8]{inputenc}
\usepackage{xcolor}
\usepackage{booktabs}
\usepackage{amsthm}
\usepackage{amsfonts}
\usepackage{listings}
\usepackage{inconsolata}
\usepackage{tikz}
\usepackage{pgfplots}
\usepackage{amsmath}
\usepackage{microtype}
\usepackage{breqn}
\usepackage{multirow}

\usepackage{subcaption}
\definecolor{pixel 0}{HTML}{FFFFFF}
\definecolor{pixel 1}{HTML}{FF0000} 

\usepackage{cleveref}

\usepackage{framed}

\newcommand{\tw}[1]{\texttt{#1}}
\newcommand{\name}{\textsc{Stevie}}

\newcommand{\popper}{\textsc{Popper}}
\newcommand{\hopper}{\textsc{Hopper}}
\newcommand{\metagol}{\textsc{Metagol}}

\theoremstyle{definition}
\newtheorem{definition}{Definition}
\newtheorem{myexample}{Example}
\newtheorem{proposition}{Proposition}
\newtheorem{lemma}{Lemma}
\newtheorem{assumption}{Assumption}

\lstnewenvironment{myalgorithm}[1][] 
{
    \lstset{ 
        basicstyle=\normalfont\ttfamily,
        showspaces=false,               
        showstringspaces=false,         
        mathescape=true,
        numbers=left,
        escapeinside={*}{*},
        columns=flexible,
        numbersep=2pt,        
        keywordstyle=\bfseries,
        keywords={,and, return, not, def, in, if, elif, else, for, foreach, while, continue,break,}
        numbers=left,
        xleftmargin=10pt,
        #1 
    }
}
{}

\makeatletter
\newcommand*\mysize{%
  \@setfontsize\mysize{8pt}{9.0}%
}
\makeatother

\usepackage{pgfkeys}
    \newenvironment{customlegend}[1][]{%
        \begingroup
        \csname pgfplots@init@cleared@structures\endcsname
        \pgfplotsset{#1}%
    }{%
        \csname pgfplots@createlegend\endcsname
        \endgroup
    }%
    \def\addlegendimage{\scriptsize\csname pgfplots@addlegendimage\endcsname}
\usepackage{comment}

\begin{document}

\maketitle


\begin{abstract}
We introduce the \emph{higher-order refactoring} problem, where the goal is to compress a logic program by discovering higher-order abstractions, such as \emph{map}, \emph{filter}, and \emph{fold}.
We implement our approach in \emph{\name{}}, which formulates the refactoring problem as a constraint optimisation problem.
Our experiments on multiple domains, including program synthesis and visual reasoning, show that refactoring can improve the learning performance of an inductive logic programming system, specifically improving predictive accuracies by 27\% and reducing learning times by 47\%.
We also show that \name{} can discover abstractions that transfer to multiple domains.
\end{abstract}
\section{Introduction}
\label{intro}
Abstraction is seen as crucial for AI \cite{saitta2013abstraction,russell:humancomp,bundy:change}.
 Despite its argued importance, 
abstraction is often overlooked in machine learning \cite{DBLP:journals/corr/abs-2002-06177,DBLP:journals/corr/abs-2102-10717}.
To address this limitation, we introduce an approach that automatically discovers \emph{higher-order} abstractions to improve the learning performance of a machine learning algorithm.

To motivate discovering higher-order abstractions, consider learning a logic program from examples to make an input string uppercase, such as \emph{[l,o,g,i,c]} $\mapsto$ \emph{[L,O,G,I,C]}.
For this problem, we could learn the program:
\[
    h_1 = 
    \begin{array}{l}
    \left\{
    \begin{array}{l}
\emph{f(A,B) $\leftarrow$ empty(A), empty(B)}\\
    \emph{f(A,B) $\leftarrow$ head(A,C), uppercase(C,E), head(B,E),}\\
\hspace{39pt} \emph{tail(A,D), f(D,F), tail(B,F)}
    \end{array}
    \right\}
    \end{array}
\]
\noindent
This program recursively uppercases each element.
Although correct, this program is verbose.
Alternatively, we could learn:
\[
    \begin{array}{l}
    \left\{
    \begin{array}{l}
\emph{f(A,B) $\leftarrow$ map(A,B,uppercase)}\\
    \end{array}
    \right\}
    \end{array}
\]
\noindent
This program uses the higher-order abstraction \emph{map} to avoid needing to learn how to recursively iterate over a list.
As this scenario shows, using abstractions can allow us to learn smaller programs, which are often easier to learn than larger ones \cite{metaho}. 

Recent work in inductive logic programming (ILP) has 
shown that using user-provided higher-order abstractions, such as \emph{map}, \emph{filter}, and \emph{fold}, can drastically improve the learning performance of an ILP system \cite{metaho,hopper}.
For instance, if given \emph{map} as input, these approaches can learn the aforementioned higher-order string transformation program.


The major limitation of these recent approaches is that they need a human to provide the necessary abstractions as input, i.e. these approaches cannot discover abstractions.

To overcome this limitation, we introduce an approach that automatically discovers useful higher-order abstractions, which can then be used by an ILP system. 
The idea is to refactor a logic program by discovering higher-order abstractions that compress it.

Our refactoring approach works in two stages: \emph{abstract} and \emph{compress}.
In the abstract stage, given a first-order program, we discover higher-order abstractions.
In the compress stage, we search for a subset of the abstractions that compresses the first-order program.

To illustrate our idea, consider the program:
\[
    h_2 = 
    \begin{array}{l}
    \left\{
    \begin{array}{l}
\emph{g(A,B) $\leftarrow$ empty(A), empty(B)}\\
    \emph{g(A,B) $\leftarrow$ head(A,C), increment(C,E), head(B,E),}\\
   \hspace{41pt} \emph{tail(A,D), g(D,F), tail(B,F)}
    \end{array}
    \right\}
    \end{array}
\]



\noindent
This program takes a list of natural numbers and adds one to each element, e.g. \emph{[3,4,5] $\mapsto$ [4,5,6]}.

Suppose we want to refactor the program $P = h_1 \cup h_2$.
In the abstract stage, we discover abstractions of $P$, such as\footnote{
There are more abstractions but we exclude them for brevity.}:
\[
    h_3 = 
    \begin{array}{l}
    \left\{
    \begin{array}{l}
\emph{ho(A,B,X) $\leftarrow$ empty(A), empty(B)}\\
    \emph{ho(A,B,X) $\leftarrow$ head(A,C), X(C,E), head(B,E),}\\
   \hspace{55pt} \emph{tail(A,D), ho(D,F,X), tail(B,F)}
    \end{array}
    \right\}
    \end{array}
\]


\noindent
The invented relation \emph{ho} defines a higher-order abstraction which corresponds to \emph{map}. The symbol \emph{X} is a higher-order variable that quantifies over predicate symbols. 

In the compress stage, we search for a subset of abstractions that compresses the input program. We formulate this problem as a \emph{constraint optimisation problem} (COP) \cite{CPHandbook}.
We output a refactored program with abstractions, such as $P' = h_3 \cup h_4$, where $h_4$ is:
\[
    h_4 = 
    \begin{array}{l}
    \left\{
    \begin{array}{l}
\emph{f(A,B) $\leftarrow$ ho(A,B,uppercase)}\\
\emph{g(A,B) $\leftarrow$ ho(A,B,increment)}
    \end{array}
    \right\}
    \end{array}
\]

\noindent
In this program, the relations \emph{f} and \emph{g} are defined with the abstraction \emph{ho}.
As this example shows, abstractions can compress a program, i.e. $P'$ has fewer literals (14) than $P$ (20).

The above scenario shows how discovering higher-order abstractions in one domain can help an ILP system perform better in that domain by allowing it to learn smaller programs.
In this paper, we show that abstractions discovered in one domain, such as program synthesis, can be reused by an ILP system in a different domain, such as chess.
Although there is much work on transfer learning \cite{torrey2009transfer} and cross-domain transfer learning \cite{DBLP:conf/icdm/KumaraswamyOKLN15}, as far as we know, we are the first to show the automatic discovery of abstractions that generalise across domains.

\subsubsection*{Novelty and Contributions}
The three main novelties of this paper are (i) the idea of discovering higher-order abstractions to refactor a logic program, (ii) encoding this refactoring problem as a COP, and (iii) showing cross-domain transfer of discovered abstractions.
The impact is that we can drastically improve the learning performance of an ILP system, compared to not discovering abstractions.
Moreover, as the idea connects many areas of AI, including machine learning, program synthesis, and constraint optimisation, we hope the idea interests a broad audience.

Overall, our contributions are:
\begin{itemize}
    \item
    We introduce the \emph{higher-order refactoring} problem, where the goal is to refactor a logic program by discovering higher-order abstractions.
    \item
    We introduce \name{} which discovers higher-order abstractions and finds an optimal solution to the higher-order refactoring problem by formulating it as a COP.
    \item
    We evaluate our approach on multiple domains, including program synthesis, visual reasoning, and robot strategy learning.
    Our empirical results show that refactoring can improve the learning performance of an ILP system, specifically improving predictive accuracies by 27\% and reducing learning times by 47\%.
    We also show that discovered abstractions can be reused across domains.
\end{itemize}
\section{Related Work}
\label{sec:related}




\textbf{Higher-order logic.}
Many authors advocate using higher-order logic to represent knowledge \cite{DBLP:conf/mi/McCarthy95,ilp20}.
Although some approaches use higher-order logic to specify the structure of learnable programs \cite{clint,mugg:metagold,hexmil}, most only learn first-order programs \cite{tilde,aleph,probfoil,dilp,metabduce,apperception,popper}.
Some approaches use higher-order abstractions \cite{metaho,hopper} but need user-defined abstractions as input.
By contrast, we automatically discover abstractions. 

\textbf{Predicate invention.}
Feng and Muggleton \shortcite{mugg:hol-inductive-generalisation} consider higher-order extensions of Plotkin's (\citeyear{plotkin:thesis}) least general generalisation, where a predicate variable replaces a predicate symbol.
By contrast, we introduce new predicate symbols, i.e. we perform \emph{predicate invention} (PI), a repeatedly stated difficult challenge \cite{cigol,pedro:pi,ilp20,russell:humancomp,kramer:ijcai20,luc:mdl,ilp30,tom:pi}.

\textbf{Representation change.}
Simon \shortcite{simon81} views abstraction as changing the representation of a problem to make it easier to solve.
Propositionalisation  \cite{nada:prop,ashwin:prop} transforms a first-order problem into a propositional one to use efficient propositional learning algorithms.
A disadvantage of propositionalisation
is the loss of a compact representation language (first-order logic).
By contrast, we change a first-order problem to a higher-order one.
Theory revision  \cite{ruth,forte,DBLP:journals/ml/PaesZC17} revises a program so that it entails missing answers or does not entail incorrect answers. 
Theory refinement improves the quality of a theory, such as its execution or readability \cite{fender,wrobel:refinement}.
By contrast, we refactor a theory to improve learning performance.

\textbf{Compression.} 
Chaitin \shortcite{chaitin:limits} emphasises compression in abstraction.
Theory compression \cite{DBLP:journals/ml/RaedtKKRT08} selects a subset of a program minimising the impact on performance with respect to the examples.
By contrast, we only consider the program, not the examples.
\textsc{Alps} \cite{seb:alps} compresses facts, while we compress logic programs.
\textsc{Knorf} \cite{knorf} refactors logic programs by framing the problem as a COP.
Whereas \textsc{Knorf} performs first-order refactoring, we perform higher-order refactoring.
Several approaches \cite{dreamcoder,stitch,babble} refactor functional programs by searching for local changes (new $\lambda$-expressions) that increase a cost function.
We differ because we (i) consider logic programs, (ii) guarantee optimal compression, and (iii) can transfer knowledge across domains.
Moreover, these approaches only evaluate the compression rate, while we show that compressing a program can improve the learning performance of an ILP system.


\section{Problem Setting}
\label{sec:setting}

We assume familiarity with logic programming \cite{lloyd:book} but have included a summary in the appendix. 
We restate key terminology.
A \emph{first-order variable} can be bound to a constant symbol or another first-order variable.
A \emph{higher-order variable} can be bound to a predicate symbol or another higher-order variable.
A \emph{clause} is a set of literals.
A clause is \emph{higher-order} if it has at least one higher-order variable.
A \emph{definite clause} is a clause with exactly one positive literal.
We use the term \emph{rule} synonymously with \emph{definite clause}.
A \emph{definite program} is a set of definite clauses with the least Herbrand model semantics.
We refer to a definite program as a \emph{logic program}.
A logic program is \emph{higher-order} if it has at least one higher-order clause.
The \emph{size(P)} of the logic program $P$ is the number of literals in $P$.
A \emph{definition} is a set of rules with the same head predicate symbol (positive predicate symbol).
The set of definitions of the logic program $P$ with the head predicate symbols $T$ is
$\delta(P) = \cup_{p \in T}{\{r \in P | \text{ the head predicate symbol of the rule $r$ is $p$}\}}
$.  
\noindent

\noindent
\subsection{Abstraction and Instantiation}
The idea of an abstraction is to replace predicate symbols with predicate variables in the body of a rule and to add these variables to the head of the rule.
We define an abstraction:

\newcommand\mydots{\makebox[1em][c]{.\hfil.\hfil.}}

\begin{definition}[\textbf{Abstraction}]
Let $P$ be a logic program,
$d \in \delta(P)$ be a definition with the head predicate symbol $h$ of arity $k$,
$\{p_1,\dots, p_n\}$ be a subset of the predicate symbols in the bodies of rules in $d$,
$x_1, \dots, x_n$ be higher-order variables,
and $h'$ be an invented predicate symbol not in $P$.
Let $a$ be the definition obtained from $d$ by replacing (1) every instance of $p_i$ with $x_i$, and (2) every literal $h(v_1, \dots, v_k)$ with the literal $h'(v_1, \dots, v_k, x_1, \dots, x_n)$.
Then $a$ is an \emph{abstraction} of $P$.
The set of all abstractions of $P$ is $\mathcal{A}(P)$.
\label{def:abstraction}
\end{definition}

\noindent
We denote invented predicate symbols with the prefix \emph{ho}.

\begin{myexample}[\textbf{Abstraction}]
\label{ex:abs}
Consider the rule:
\[
\begin{tabular}{l}
    \emph{f(A) $\leftarrow$ head(A,B), one(B), tail(A,C), head(C,D), one(D)}
    \end{tabular}
\]
Some abstractions of this rule are:
\[
\begin{tabular}{l}
 \emph{ho$_1$(A,X) $\leftarrow$ X(A,B), one(B), tail(A,C), X(C,D), one(D)}\\
 \emph{ho$_2$(A,X) $\leftarrow$ head(A,B), X(B), tail(A,C), head(C,D), X(D)}\\
 \emph{ho$_3$(A,X,Y) $\leftarrow$ X(A,B), Y(B), tail(A,C), X(C,D), Y(D)}
\end{tabular}
\]
Consider the recursive definition:
\begin{center}
\begin{tabular}{l}
\emph{g(A,B) $\leftarrow$ head(A,B)}\\
\emph{g(A,B) $\leftarrow$ tail(A,C), g(C,B)}
\end{tabular}
\end{center}
Some abstractions of this definition are:
\[ 
    \begin{tabular}{l}
\emph{ho$_4$(A,B,X) $\leftarrow$ X(A,B)}\\
\emph{ho$_4$(A,B,X) $\leftarrow$ tail(A,C), ho$_4$(C,B,X)}\\ [0.15cm]
\emph{ho$_5$(A,B,X) $\leftarrow$ head(A,B)}\\
\emph{ho$_5$(A,B,X) $\leftarrow$ X(A,C), ho$_5$(C,B,X)}\\ [0.15cm]
\emph{ho$_6$(A,B,X,Y) $\leftarrow$ X(A,B)}\\
\emph{ho$_6$(A,B,X,Y) $\leftarrow$ Y(A,C), ho$_6$(C,B,X,Y)}
    \end{tabular}
\] 
 
 
\end{myexample}

\noindent

\noindent
An instantiation replaces predicate variables in an abstraction with predicate symbols:
\begin{definition}[\textbf{Instantiation}]
Let 
$P$ be a logic program,
$h(v_1, \dots, v_k)$ be a head literal in $P$,
$h'(v_1, \dots, v_k, x_1, \dots, x_n)$ be a head literal in $\mathcal{A}(P)$,
$x_1, \dots, x_n$ be higher-order variables,
and $p_1, \dots, p_n$ be predicate symbols in the bodies of rules in $P$.
Then the rule $h(v_1, ..., v_k) \leftarrow h'(v_1, ..., v_k, p_1, \dots, p_n)$ is an \emph{instantiation}.
The set of all instantiations of abstractions of $P$ is $\mathcal{I}(\mathcal{A}(P))$.
\label{def:instantiation}
\end{definition}

\begin{myexample}[\textbf{Instantiation}]
Some instantiations of the abstractions in Example \ref{ex:abs} are:
\begin{center}
\begin{tabular}{l}
\emph{f(A) $\leftarrow$ ho$_2$(A,one)}\\
\emph{f(A) $\leftarrow$ ho$_3$(A,head,one)}\\
\emph{g(A,B) $\leftarrow$ ho$_6$(A,B,head,tail)}
\end{tabular}
\end{center}
\end{myexample}

\noindent

\noindent

\noindent


\subsection{Higher-Order Refactoring Problem}
When we refactor a program, we want to preserve its semantics.
However, we only need to preserve the semantics with respect to head predicate symbols.
Therefore, we reason about the least Herbrand model restricted to a set of predicate symbols:
\begin{definition}[\textbf{Restricted least Herbrand model}]
Let $P$ be a logic program,
$M(P)$ be the least Herbrand model of $P$,
and $T$ be the head predicate symbols of $P$.
Then the least Herbrand model of $P$ restricted to $T$ is $M_T(P) = \{ a \in M(P) \,|\, \text{the predicate symbol of $a$ is in } T\}$.
\end{definition}

\noindent
We define the \emph{higher-order refactoring} problem:

\begin{definition}[\textbf{Higher-order refactoring problem}]
\label{def:prob}
Let $P$ be a logic program
and $T$ be the head predicate symbols of $P$.
Then the \emph{higher-order refactoring problem} is to find $Q \subseteq P \cup \mathcal{A}(P) \cup \mathcal{I}(\mathcal{A}(P))$ such that $M_T(Q) == M_T(P)$.
We call $Q$ a \emph{solution} to the refactoring problem.
\end{definition}
\begin{myexample}[\textbf{Refactoring}]
A refactoring of the program $P$ in Section \ref{intro} is $P'$.
\end{myexample}


\noindent
Our goal is to perform \emph{optimal refactoring}:

\begin{definition}[\textbf{Optimal refactoring}]\label{def:optprob}
Let $P$ be a logic program, $T$ be the head predicate symbols of $P$, and $cost$ be a function which maps logic programs to integers.
Then $Q$ is an \emph{optimal} solution when (i) $Q$ is a solution to the refactoring problem, and (ii) there is no $Q' \subseteq  P \cup \mathcal{A}(P) \cup \mathcal{I}(\mathcal{A}(P))$ such that $Q'$ is a solution to the refactoring problem and $cost(Q') < cost(Q)$.
\end{definition}



    \noindent
In the next section, we introduce \name{}, which finds an optimal solution to the refactoring problem.

\section{\name{}}
\label{sec:impl}
Algorithm \ref{alg:stevie} shows our \name{} algorithm, which works in two stages: \emph{abstract} and \emph{compress}.
In the \emph{abstract} stage, given a first-order logic program, \name{} builds abstractions and instantiations.
In the \emph{compress} stage, \name{} searches for a subset of the abstractions and instantiations which compresses the input program.
\name{} formulates this search problem as a COP.
We describe these two stages in turn.
The appendix includes an example of refactoring.

\begin{algorithm}[ht]
\footnotesize
{
\begin{myalgorithm}[]
def stevie(P, k):
  abstractions, instantiations = abstract(P, k)
  return compress(P, abstractions, instantiations)

def abstract(P, k):
  abstractions, instantiations = {}, {}
  for d in $\delta(P)$: 
    for size in 1 to k:
      for $\psi$ in subsets(nonrecbodypreds(d), size):
        abs, inst = create_abs_inst(d, $\psi$)
        if equivalent(abs, abstractions):
          inst = redefine(inst, abs, abstractions)
        else:
          abstractions += abs 
        instantiations += {inst}
  return abstractions, instantiations
\end{myalgorithm}
\caption{
\name{}
}
\label{alg:stevie}
}
\end{algorithm}
\noindent


\subsection{Abstract}
\label{sec:abstract}
In the \emph{abstract} stage (line 2), \name{} builds abstractions and instantiations. 
To build abstractions for the logic program $P$, 
for each definition $d \in \delta(P)$ and subset $\psi$ of at most $k$ predicate symbols in the bodies of rules in $d$,
\name{} calls the function \emph{create\_abs\_inst(d, $\psi$)} (line 10).
The value $k$ is a user parameter.
This function follows Definition \ref{def:abstraction} and  replaces every $p_i \in \psi$ in $d$ with a new higher-order variable $x_i$, adds each $x_i$ to the arguments of the literals with the predicate symbol $h$, where $h$ is the head predicate symbol of $d$, and replaces every occurrence of $h$ with an invented predicate symbol $h'$.
For instance, if $d$ is the rule in Example \ref{ex:abs} and $\psi$=\emph{\{head, one\}}, the function replaces \emph{head} with \emph{X} and \emph{one} with \emph{Y} to build the abstraction \emph{ho$_3$} in Example  \ref{ex:abs}.
\name{} never abstracts recursive predicate symbols (line 9) as this would change the semantics.
This function also returns an instantiation (Definition \ref{def:instantiation}) by replacing predicate variables in an abstraction with $\psi$.
\name{} prunes abstractions that are identical up to renaming of their head predicate symbol (line 11).
In such cases, \name{} redefines the instantiation in terms of the existing equivalent abstraction (line 12). For instance, consider the rules:
\begin{center}
\begin{tabular}{l}
\emph{f$_1$(A) $\leftarrow$ head(A,B), one(B)}\\
\emph{f$_2$(A) $\leftarrow$ head(A,B), two(B)}
\end{tabular}
\end{center}
The abstractions of the $f_1$ and $f_2$ rules with $\psi=\emph{\{one\}}$ and $\psi=\emph{\{two\}}$ respectively are equivalent up to renaming of the head predicate symbols, i.e. both of these rules have the abstraction \emph{ho(A, X) $\leftarrow$ head(A,B), X(B)}.

\subsection{Compress}
\label{compress}

In the \emph{compress} stage, \name{} searches for a subset of abstractions and instantiations that compresses the input program (line 3).
\name{} formulates this search problem as a COP.
Given (i) a set of decision variables, (ii) a set of constraints, and (iii) an objective function, a COP solver finds an assignment to the decision variables that satisfies all the specified constraints and minimises the objective function.
 
We describe our COP encoding. 
We assume an input logic program $P$. 

\subsubsection{Decision Variables}
\name{} uses three types of decision variables. 
First, for each definition $d \in \delta(P)$ and abstraction $a \in \mathcal{A}(P)$, we use a
Boolean variable $i^d_a$ to indicate whether an instantiation of $a$ defining $d$ is selected.
We later use these variables to 
ensure that a definition is defined with at most one instantiation. 
Second, for each definition $d \in \delta(P)$, we use a Boolean variable $n_d$ to indicate 
that no instantiation has been selected for $d$. 
These variables allow \name{} to not introduce abstractions and instantiations if they overall increase the complexity of the refactored program. 
Third, for each abstraction $a \in \mathcal{A}(P)$, we use a
Boolean variable $s_a$ to indicate that at least one instantiation of $a$ is selected.
\name{} uses these variables to determine the size of the refactored program.

\subsubsection{Constraints}
\name{} imposes two types of constraints.
First, for each definition $d \in \delta(P)$, \name{} uses a constraint to ensure that at most one instantiation is selected for $d$: 
$$\left ( \sum_{a \in \mathcal{A}(P)}i^d_a \right ) + n_d = 1$$
This constraint is necessary to identify definitions which are not refactored.

Second, for each abstraction $a \in \mathcal{A}(P)$, \name{} uses a constraint to ensure that the 
variable $s_a$ is true if and only if an instantiation of $a$ is used to refactor at least one definition\footnote{The OR-tools solver that we use treats Boolean variables as integer variables with domain $\{0, 1\}$. Therefore, both
arithmetic and Boolean operators apply to them.}:
$$s_a \leftrightarrow \bigvee_{d \in \delta(P)} i^{d}_a$$

\subsubsection{Objective}
Our objective function is the summation of three components: (1) the size of non-abstracted definitions,
(2) the size of selected abstractions and instantiations, 
and (3) a penalty on the number of higher-order variables.
We describe these in turn.

The size of non-abstracted definitions is:
\begin{dmath}
\sum_{d \in \delta(P)} size(d) \times n_d
\label{c1}
\end{dmath}
\noindent
An instantiation is a rule with one body literal so has size 2. 
The size of selected abstractions and instantiations is:
\begin{dmath}
\underbrace{\sum_{a \in \mathcal{A}(P)} size(a)\times s_a}_{\text{selected abstractions}}+ \underbrace{\sum_{d \in \delta(P), a \in \mathcal{A}(P)} 2 \times i_a^d}_{\text{selected instantiations}}
\end{dmath}


\noindent
\name{} penalises the number of higher-order variables in a refactoring. 
Without it, \name{} often selects abstractions that remove all the predicate symbols in a definition.
For instance, \name{} might introduce abstractions such as:
\begin{center}
\begin{tabular}{l}
\emph{ho(A,B,X,Y,Z) $\leftarrow$ X(A,C), Y(C,D), Z(D,B)} 
\end{tabular}
\end{center}
Therefore, 
\name{} uses the following penalty, where \emph{ho\_vars(a)} is the number of higher-order variables in the abstraction $a$:
\begin{dmath}
\sum_{a \in \mathcal{A}(P)} ho\_vars(a) \times s_a \label{c4}
\end{dmath}
 \noindent
As we show in our experiments, this penalty allows us to find abstractions that lead to better learning performance.
\subsection{Correctness}
We prove the correctness of \name{}:
\begin{theorem} \name{} solves the optimal refactoring problem with respect to our objective function.
\end{theorem}
\noindent
The proof is in the appendix. 
To show this result, we show that 
(i) \name{} generates all abstractions and instantiations (Definitions \ref{def:abstraction} and \ref{def:instantiation}),
(ii) any solution to the encoding is a solution to the higher-order refactoring problem (Definition \ref{def:prob}), 
and (iii) the solver finds an optimal solution
(Definition \ref{def:optprob}) with respect to our objective function.
\section{Experiments}
\label{sec:exp}
To test our claim that higher-order refactoring can improve the performance of an ILP system, our experiments aim to answer the question:
\begin{description}
\item[Q1] Can higher-order refactoring improve predictive accuracies and reduce learning times? 
\end{description}
To answer \textbf{Q1}, we compare the performance of an ILP system with and without the ability to use abstractions discovered by \name{}.
We use the ILP system \hopper{} \cite{hopper} because it can learn recursive programs, perform predicate invention, and use higher-order abstractions as BK\footnote{We also considered \metagol{}$_{HO}$ \cite{metaho} but it needs user-provided metarules which are difficult to obtain \cite{ilp30}.}. 


To understand the impact of penalising the number of higher-order variables (component (3) in Section \ref{compress}),
our experiments aim to answer the question:
\begin{description}
    \item[Q2] 
    What is the impact of penalising the number of higher-order variables on learning performance?
\end{description}
To answer \textbf{Q2}, we compare \name{} with and without the penalty on the number of higher-order variables.

To understand the scalability of our approach, our experiments aim to answer the question:
\begin{description}
    \item[Q3] How long does \name{} take given larger programs?
\end{description}
To answer \textbf{Q3}, we measure the refactoring time of \name{} on progressively larger programs.

\pgfplotstableread{
id    xs         av       sem
0    0  71.647045  1.666442
1   30  72.223182  1.664352
2   60  77.736591  1.665023
3   90  81.766818  1.609635
4  120  89.310682  1.370023
5  132  90.872273  1.283075
}\stevieacc

\pgfplotstableread{
id    xs         av       sem
0    0	69.23818181818180	1.630876603137150
1   30	64.02613636363640	1.4895964961613900
2   60	69.04159090909090	1.619905283882410
3   90	79.15636363636360	1.6479308956504000
4  120	82.30409090909090	1.6100449158111900
5  132	84.09636363636360	1.5574429094405500
}\stevieaccnopenalty

\pgfplotstableread{
id    xs         av       sem
0    0  72.194545  1.673833
1   30  70.720455  1.658542
2   60  71.658182  1.666253
3   90  71.992500  1.673635
4  120  71.672727  1.666251
5  132  71.241364  1.663964
}\prologacc

\pgfplotstableread{
id    xs         av       sem
0    0  71.485227  1.659696
1   30  71.267045  1.663667
2   60  71.120000  1.658090
3   90  70.659318  1.655520
4  120  71.466591  1.665241
5  132  71.248409  1.663175
}\norefacc

\pgfplotstableread{
id    xs          av        sem
0    0  412.148058  28.025583
1   30  466.814388  28.767303
2   60  453.128168  28.198571
3   90  367.029655  27.544442
4  120  244.320034  24.147140
5  132  218.380208  22.530982
}\stevietime

\pgfplotstableread{
id    xs          av        sem
0    0	466.6767211192910	28.093161529198100
1   30	638.8466812454640	25.96671434448600
2   60	574.356244961891	27.083169695162800
3   90	432.00769516339500	27.223108333756000
4  120	378.77489662281400	27.101301472631500
5  132	348.7183128084140	26.371482635538600
}\stevietimenopenalty

\pgfplotstableread{
id    xs          av        sem
0    0  550.030784  27.506837
1   30  564.870481  27.630364
2   60  551.146870  27.655984
3   90  542.450528  27.746104
4  120  547.352189  27.735564
5  132  550.284409  27.945117
}\prologtime

\pgfplotstableread{
id    xs          av        sem
0    0  449.860674  28.103501
1   30  458.543153  27.911401
2   60  440.934142  28.250164
3   90  459.671120  28.091202
4  120  454.726787  28.144943
5  132  459.480814  28.303937
}\noreftime

\pgfplotstableread{
id    xs        av       sem
0    0  8.250000  0.219149
1   30  6.666667  0.423612
2   60  4.228346  0.375175
3   90  4.020690  0.312184
4  120  3.685393  0.240504
5  132  3.686486  0.273583
}\stevielength

\pgfplotstableread{
id    xs        av       sem
0   0	8.779069767441860	0.2569957777180330
1   30	6.591549295774650	0.6836926206363270
2   60	6.0	0.7268732153558930
3   90	4.598540145985400	0.47154540857717500
4  120	3.3793103448275900	0.2559762631021990
5  132	3.490322580645160	0.23575774170139000
}\stevielengthnopenalty

\pgfplotstableread{
id    xs        av       sem
0    0  4.285714  0.172846
1   30  4.195652  0.176389
2   60  4.260417  0.178471
3   90  4.247423  0.171555
4  120  4.250000  0.173963
5  132  4.319149  0.225980
}\prologlength

\pgfplotstableread{
id    xs        av       sem
0    0  8.625000  0.246132
1   30  8.873684  0.338709
2   60  8.542553  0.244203
3   90  8.989247  0.472105
4  120  8.568421  0.239752
5  132  8.521277  0.245168
}\noreflength

\pgfplotstableread{
id      xs          av         sem
0  112.0    0.065394    0.010739
1  241.6    0.619361    0.264287
2  327.2   57.626006   54.852515
3  437.4  318.256017  151.265914
4  460.4  958.364692  662.147769
}\resstevie

\pgfplotstableread{
id      xs          av         sem
0  106.2    0.08118602560123310	0.022411521015584500
1  212.4    0.12432707299755500	0.01577994880057890
2  327.6   0.8825246855981340	0.41411283728487000
3  432.2  59.89870540860070	51.325422757708200
4  475.0  92.44550210239890	50.01005072729410
}\resstevienopenalty

\definecolor{mygreen}{cmyk}{0.9,0,1.0,0.05}
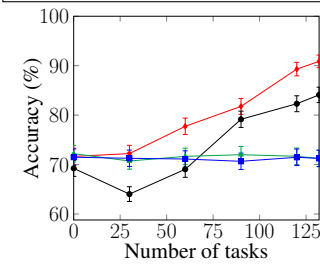
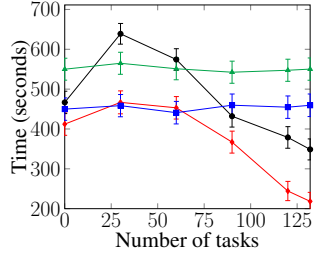
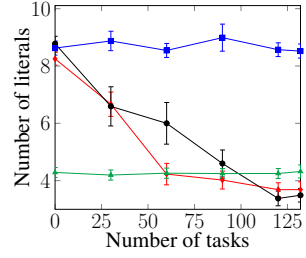
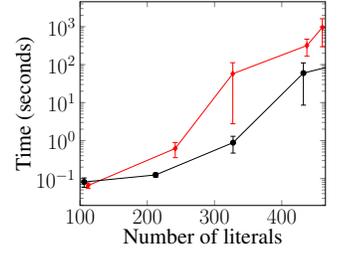
\begin{figure*}[h!]
  \begin{minipage}{0.05\textwidth}
      \centering
\begin{tikzpicture}
    \centering
\begin{customlegend}[legend columns=5,legend style={nodes={scale=1, transform shape},align=left,column sep=0ex,font=\small},
        legend entries={\name, \name$_{\text{no penalty}}$, \emph{no refactoring}, \emph{Prolog library}}]
        \addlegendimage{red,mark=diamond*}
        \addlegendimage{black,mark=oplus*}
        \addlegendimage{blue,mark=square*}
        \addlegendimage{mygreen,mark=triangle*}
\end{customlegend}
\end{tikzpicture}
\end{minipage}\hfill\\
\resizebox{2.1\columnwidth}{!}{
  \begin{minipage}{0.5\columnwidth}
 \subfloat[Predictive accuracy versus the number of tasks. \label{fig:accsynthesis}]{
\begin{tikzpicture}[scale=0.465]
\begin{axis}[
  legend style={at={(0.5,0.35)},anchor=west},
   legend style={font=\normalsize},
    label style={font=\huge},
    tick label style={font=\huge},
   xtick={0,25,50,75,100,125,150},
   ytick={60,70,80,90,100},
   ylabel style={yshift=-1mm},
  xlabel=Number of tasks,
  ylabel=Accuracy (\%),
  xmin=0,
  xmax=132,
  ymax=100,
  ]
    \addplot[red,mark=diamond*,
                error bars/.cd,
                y dir=both,
                error mark,
                y explicit]table[x=xs,y=av, y error=sem] {\stevieacc};
    \addplot[black,mark=oplus*,
                error bars/.cd,
                y dir=both,
                error mark,
                y explicit]table[x=xs,y=av, y error=sem] {\stevieaccnopenalty};
    \addplot[mygreen,mark=triangle*,
                error bars/.cd,
                y dir=both,
                error mark,
                y explicit]table[x=xs,y=av, y error=sem] {\prologacc};
\addplot[blue,mark=square*,
                error bars/.cd,
                y dir=both,
                error mark,
                y explicit]table[x=xs,y=av, y error=sem] {\norefacc};

\end{axis}
\end{tikzpicture}
}
\end{minipage}
\hfill
  \begin{minipage}{0.5\columnwidth}
 \subfloat[Learning time versus the number of tasks.
\label{fig:timesynthesis}]{
\begin{tikzpicture}[scale=0.465]
\begin{axis}[
  legend style={at={(0.5,0.35)},anchor=west},
   legend style={font=\normalsize},
    label style={font=\huge},
    tick label style={font=\huge},
    ylabel style={yshift=0mm},
   xtick={0,25,50,75,100,125,150},
  ytick={0,100,200,300,400,500,600,700},
  xlabel=Number of tasks,
  ylabel=Time (seconds),
  xmin=0,
  xmax=132,
  ymin=200,
  ]
      \addplot[red,mark=diamond*,
                error bars/.cd,
                y dir=both,
                error mark,
                y explicit]table[x=xs,y=av, y error=sem] {\stevietime};
      \addplot[black,mark=oplus*,
                error bars/.cd,
                y dir=both,
                error mark,
                y explicit]table[x=xs,y=av, y error=sem] {\stevietimenopenalty};
                stevietimenopenalty
    \addplot[mygreen,mark=triangle*,
                error bars/.cd,
                y dir=both,
                error mark,
                y explicit]table[x=xs,y=av, y error=sem] {\prologtime};
\addplot[blue,mark=square*,
                error bars/.cd,
                y dir=both,
                error mark,
                y explicit]table[x=xs,y=av, y error=sem] {\noreftime};

\end{axis}
\end{tikzpicture}
}
\end{minipage}
\hfill
  \begin{minipage}{0.5\columnwidth}
 \subfloat[Average solution length versus the number of tasks.\label{fig:literals}]{
\begin{tikzpicture}[scale=0.465]
\begin{axis}[
  legend style={at={(0.5,0.35)},anchor=west},
   legend style={font=\normalsize},
    label style={font=\huge},
    tick label style={font=\huge},
   xtick={0,25,50,75,100,125,150},
    ylabel style={yshift=-3mm},
  xlabel=Number of tasks,
  ylabel=Number of literals,
  xmin=0,
  xmax=132,
  ymin=3,
  ]
    \addplot[red,mark=diamond*,
                error bars/.cd,
                y dir=both,
                error mark,
                y explicit]table[x=xs,y=av, y error=sem] {\stevielength};
    \addplot[black,mark=oplus*,
                error bars/.cd,
                y dir=both,
                error mark,
                y explicit]table[x=xs,y=av, y error=sem] {\stevielengthnopenalty};
    \addplot[mygreen,mark=triangle*,
                error bars/.cd,
                y dir=both,
                error mark,
                y explicit]table[x=xs,y=av, y error=sem] {\prologlength};
\addplot[blue,mark=square*,
                error bars/.cd,
                y dir=both,
                error mark,
                y explicit]table[x=xs,y=av, y error=sem] {\noreflength};

\end{axis}
\end{tikzpicture}
}
\end{minipage}
\hfill
  \begin{minipage}{0.5\columnwidth}
 \subfloat[Optimal refactoring time versus the program size.
\label{fig:stevie}]{
\centering
\begin{tikzpicture}[scale=0.465]
    \begin{axis}[
    scaled x ticks = false,
    xlabel=Number of literals,
    ylabel=Time (seconds),
    xtick={100,200,300,400,500},    
    ylabel style={yshift=-2mm},
    xmin=100,
    xmax=465,    
    ymode=log,
    ylabel style={yshift=5mm},
    label style={font=\huge},
    tick label style={font=\huge},
    legend style={legend pos=north west,style={nodes={right}}}
    ]
        \addplot[red,mark=diamond*,
                error bars/.cd,
                y dir=both,
                error mark,
                y explicit]table[x=xs,y=av, y error=sem] {\resstevie};
        \addplot[black,mark=oplus*,
                error bars/.cd,
                y dir=both,
                error mark,
                y explicit]table[x=xs,y=av, y error=sem] {\resstevienopenalty};
    \end{axis}
  \end{tikzpicture}
  }
\end{minipage}
}
\caption{Results for the \emph{program synthesis} domain.
\label{programsynthesis}
}
\end{figure*}


To test our claim that abstractions discovered in one domain can be reused in different domains, our experiments aim to answer the question:
\begin{description}
\item[Q4] Can higher-order refactoring improve performance across domains?
\end{description}
\noindent
To answer \textbf{Q4}, we compare the performance of \hopper{} with and without abstractions discovered by \name{} in a different domain.

\paragraph{Settings.}
\hopper{} uses types to restrict the hypothesis space (the set of all programs). 
We use a bottom-up procedure to infer types for the abstractions discovered by \name{} from the types of the first-order BK.
We set \hopper{} to use at most three abstractions in a program.
We allow \hopper{} to use three threads.
We use SWI-Prolog to execute the programs learned by \name{} and \hopper{}.
We allow \name{} to discover abstractions with at most three higher-order variables.
\name{} uses the CP-SAT solver \cite{ortools}. 
We use a c6a AWS instance with 32vCPU and 64GB of memory. 
\name{} uses a single CPU. 


\paragraph{Method.} 
We measure the predictive accuracy (the proportion of correct predictions on test data) and learning time of \hopper{}. We use a maximum learning time of 15 minutes per task and return the best solution found by \hopper{} in this time limit.
We use a timeout of 1 hour for \name{} and return the best refactoring found in this time limit.
We repeat all the experiments 5 times and calculate the mean and standard error. The error bars in the figures and tables denote standard error.
We rename the abstractions in the figures for clarity.


\subsection{Q1: Learning Performance} 
\label{exp1}


\paragraph{Domain.}

We use a dataset of 176 program synthesis tasks and reserve 25\% as held-out tasks.
The tasks are designed to use a variety of higher-order constructs and require learning recursive programs.
 For instance, the dataset includes the tasks
\emph{counteven}, \emph{filterodd} (Figure \ref{fig:filterodd}), 
and
\emph{maxlist} (Figure \ref{fig:maxlist}).
The appendix contains more details, such as example solutions.



\paragraph{Method.}
Our method has three steps.
In step 1, we use \hopper{} to independently learn solutions for $n$ tasks.
In step 2, we use \name{} to refactor the programs learned in step 1.
In step 3, we add the abstractions discovered in step 2 by \name{} to the BK of \hopper{}. 
We then use \hopper{} on the held-out tasks.
We vary the number $n$ of tasks in step 1 and measure the performance of \hopper{} in step 3.
The baseline (\emph{no refactoring}) is when we do not use \name{} in step 2, i.e. the baseline is \hopper{} without the abstractions discovered by \name{}.
As a second baseline, we use seven standard higher-order abstractions (\emph{maplist}, \emph{foldl}, \emph{scanl}, \emph{convlist}, \emph{partition}, \emph{include}, and \emph{exclude}) from the SWI-Prolog library \emph{apply}\footnote{\url{https://www.swi-prolog.org/pldoc/man?section=apply}}.
The appendix includes a description of these abstractions.


\subsubsection{Results}
\setlength{\OuterFrameSep}{-5pt}
 \begin{figure}[ht!]
 \setlength{\fboxsep}{4pt}
\fbox{
\begin{minipage}{0.46\textwidth}
 \subfloat[\emph{filterodd} program which removes the odd elements of a list.\label{fig:filterodd}]{
\mysize
\begin{tabular}{p{0.95\textwidth}}
\texttt{filterodd(A,B) $\leftarrow$ empty(A),empty(B)} \\
\texttt{filterodd(A,B) $\leftarrow$ head(A,C),tail(A,D),odd(C),}\\
\hspace{72pt}\texttt{filterodd(D,B)} \\
\texttt{filterodd(A,B) $\leftarrow$ head(A,C),tail(A,D),even(C),}\\
\hspace{72pt}\texttt{filterodd(D,E),head(B,C),tail(B,E)} \\
\end{tabular}
}
\end{minipage}
}
\fbox{
\begin{minipage}{0.46\textwidth}
 \subfloat[\emph{filterpos} program which removes positive elements of a list. \label{fig:filterpositive}]{
\mysize
\begin{tabular}{p{0.95\textwidth}}
\texttt{filterpos(A,B) $\leftarrow$ empty(A),empty(B)} \\
\texttt{filterpos(A,B) $\leftarrow$ head(A,C),tail(A,D),pos(C),}\\
\hspace{71pt} \texttt{filterpos(D,B)} \\
\texttt{filterpos(A,B) $\leftarrow$ head(A,C),tail(A,D),neg(C),}\\
\hspace{72pt}\texttt{filterpos(D,E),head(B,C),tail(B,E)}\\
\end{tabular}
}
\end{minipage}
}
\fbox{
\begin{minipage}{0.46\textwidth}
 \subfloat[
 Higher-order \emph{ho$\_$filter} abstraction discovered by \name{} which returns elements of a list where $Y$ holds and $X$ does not.\label{fig:filter}]{
  \mysize
\begin{tabular}{p{0.95\textwidth}}
\texttt{ho_filter(A,B,X,Y) $\leftarrow$ empty(A),empty(B)} \\
\texttt{ho_filter(A,B,X,Y) $\leftarrow$ head(A,C),tail(A,D),X(C),}\\
\hspace{87pt}\texttt{ho_filter(D,B,X,Y)} \\
\texttt{ho_filter(A,B,X,Y) $\leftarrow$ head(A,C),tail(A,D),Y(C),head(B,C),}\\
\hspace{87pt}\texttt{ho_filter(D,E,X,Y),tail(B,E)}\\
\end{tabular}
}
\end{minipage}
}
\fbox{
\begin{minipage}{0.46\textwidth}
 \subfloat[Instantiations.\label{fig:filtercompressed}]{
  \mysize
\begin{tabular}{p{0.95\textwidth}}
\texttt{filterodd(A,B) $\leftarrow$ \emph{ho_filter}(A,B,odd,even)} \\
\texttt{filterpos(A,B) $\leftarrow$ \emph{ho_filter}(A,B,pos,neg)} \\
\end{tabular}
}
\end{minipage}
}
\caption{Example of \name{} discovering the higher-order abstraction \emph{ho$\_$filter} to compress programs.
}
\end{figure}
 \begin{figure}[ht!]
 \setlength{\fboxsep}{4pt}
\fbox{
\begin{minipage}{0.46\textwidth}
 \subfloat[\emph{multlist} program which returns the cumulative product of the elements of a list. \label{fig:multlist}]{
\mysize
\begin{tabular}{p{0.95\textwidth}}
\texttt{multlist(A,B) $\leftarrow$ empty(A),one(B).} \\
\texttt{multlist(A,B) $\leftarrow$ head(A,C),tail(A,D),}\\
\hspace{69pt}\texttt{multlist(D,E),mult(C,E,B)}\\
\end{tabular}}
\end{minipage}
}
\fbox{
\begin{minipage}{0.46\textwidth}
 \subfloat[\emph{maxlist} program which returns the maximum element of a list.\label{fig:maxlist}]{
\mysize
\begin{tabular}{p{0.95\textwidth}}
\texttt{maxlist(A,B) $\leftarrow$ empty(A),zero(B).} \\
\texttt{maxlist(A,B) $\leftarrow$ head(A,C),tail(A,D),}\\
\hspace{65pt}\texttt{maxlist(D,E),max(C,E,B)}\\
\end{tabular}}
\end{minipage}
}
\fbox{
\begin{minipage}{0.46\textwidth}
 \subfloat[Higher-order \emph{ho$\_$fold} abstraction discovered by \name{} which recursively combines all elements of a list using the higher-order predicate $X$ and the default value returned by $Y$.\label{fig:fold}]{
  \mysize
\begin{tabular}{p{0.95\textwidth}}
\texttt{ho_fold(A,B,X,Y) $\leftarrow$ empty(A),X(B)} \\
\texttt{ho_fold(A,B,X,Y) $\leftarrow$ head(A,C),tail(A,D),}\\
\hspace{78pt} \texttt{ho_fold(D,E,X,Y),Y(C,E,B)} \\
\end{tabular}
}
\end{minipage}
}
\fbox{
\begin{minipage}{0.46\textwidth}
 \subfloat[Instantiations.\label{fig:compressed}]{
  \mysize
\begin{tabular}{p{0.95\textwidth}}
\texttt{multlist(A,B) $\leftarrow$ \emph{ho_fold}(A,B,one,mult)} \\
\texttt{maxlist(A,B) $\leftarrow$ \emph{ho_fold}(A,B,zero,max)} \\
\end{tabular}
}
\end{minipage}
}
\fbox{
\begin{minipage}{0.46\textwidth}
 \subfloat[\emph{sumlistplus3} program. \label{fig:sumlistplus3fo}]{
\mysize
\begin{tabular}{p{0.95\textwidth}}
\texttt{sumlistplus3(A,B) $\leftarrow$ empty(A),one(C),succ(C,D),succ(D,B)}\\
\texttt{sumlistplus3(A,B) $\leftarrow$ head(A,C),tail(A,D),}\\
\hspace{82pt} \texttt{sumlistplus3(D,E),sum(C,E,B)}\\
\end{tabular}}
\end{minipage}
}
\fbox{
\begin{minipage}{0.46\textwidth}
\subfloat[\emph{sumlistplus3} program using the abstraction \emph{ho$\_$fold}. 
The predicate \emph{inv} is invented by \hopper{}.\label{fig:sumlistplus3ho}]{
 \mysize
\begin{tabular}{p{0.95\textwidth}}
\texttt{sumlistplus3(A,B) $\leftarrow$ \emph{ho_fold}(A,B,inv,sum)} \\
\texttt{inv(A) $\leftarrow$ one(B),succ(B,C),succ(C,A)}
\end{tabular}}
\end{minipage}
}
\caption{Example of \name{} discovering the higher-order abstraction \emph{ho$\_$fold} to compress programs.
}
\end{figure}

Figure \ref{fig:accsynthesis} shows that our approach (\name) can increase predictive accuracies by 27\% compared to the baselines. 
Figure \ref{fig:timesynthesis} shows that our approach can reduce learning times by 47\% compared to the baselines.
A chi-square test and a Mann-Whitney U rank test confirm ($p<0.01$) the significance of the difference in accuracy and learning times respectively.

To illustrate higher-order refactoring, 
consider the tasks \emph{filterodd} and \emph{filterpos}. Figures \ref{fig:filterodd} and \ref{fig:filterpositive} show the programs learned by \hopper{} for these tasks. 
\name{} compresses these programs by discovering the abstraction shown in Figure \ref{fig:filter}. 
This abstraction keeps elements in a list where the higher-order predicate \emph{Y} holds and removes elements where the higher-order predicate \emph{X} holds, i.e. this abstraction filters a list.
\name{} thus compresses the program from 30 literals (Figures \ref{fig:filterodd} and \ref{fig:filterpositive}) to 19 literals (Figures \ref{fig:filter} and \ref{fig:filtercompressed}).

As a second illustration, consider the tasks \emph{multlist} (Figure \ref{fig:multlist}) and \emph{maxlist} (Figure \ref{fig:maxlist}). 
\name{} compresses these programs by discovering the abstraction \emph{fold} (Figure \ref{fig:fold}). 
This abstraction recursively combines the elements of a list using the higher-order predicate $X$ and the default value given by the higher-order predicate $Y$.
\name{} thus compresses the program from 16 literals (Figures \ref{fig:multlist} and \ref{fig:maxlist}) to 12 (Figures \ref{fig:fold} and \ref{fig:compressed}). 
Moreover, \hopper{} reuses the abstraction \emph{fold} to learn programs for more complex tasks.
For instance, without abstraction, \hopper{} learns a program for \emph{sumlistplus3} with 10 literals (Figure \ref{fig:sumlistplus3fo}), whereas with the abstraction \emph{fold} it learns a solution with only 6 literals (Figure \ref{fig:sumlistplus3ho}).

\name{} can discover many abstractions, such as \emph{map}, \emph{count}, \emph{iterate}, \emph{until}, \emph{member}, and \emph{all}. 
The appendix includes all the abstractions discovered by \name{}.
\hopper{} can combine these abstractions to learn succinct programs for complex tasks.
For instance, for the task \emph{sumunicodes}, \hopper{} learns a compact solution (1 rule and 3 literals) which uses the abstractions \emph{map} and \emph{fold} (Figure \ref{fig:sumunicodes}).
Without abstractions, \hopper{} would need to learn a program with at least 5 rules and 21 literals.

Figure \ref{fig:literals} shows that refactoring typically reduces the size of programs learned by \hopper{} from 8 to 4 literals. 
As recent work shows \cite{metaho,hopper}, learning smaller programs can improve learning performance since the system searches a smaller hypothesis space.

Overall, these results suggest that higher-order refactoring can substantially improve learning performance (\textbf{Q1}).


\subsection{Q2: Higher-Order Variables Penalty}
Figures \ref{fig:accsynthesis} and \ref{fig:timesynthesis} show that penalising the number of higher-order variables can increase predictive accuracies by 8\% and decrease learning times by 37\%.
A chi-square test and a Mann-Whitney U rank test confirm ($p<0.01$) the significance of the difference in accuracy and learning times respectively.
This result suggests that component (3) of our objective function can help improve performance. 
Without the penalty, \name{} finds abstractions with many higher-order variables. 
These abstractions are less helpful as \hopper{} must search through the space of all possible instantiations which is larger when there are more higher-order variables.
This result indicates that not all abstractions can help and that finding good abstractions is important.
Overall, these results suggest penalising the number of higher-order variables can improve learning performance (\textbf{Q2}).

 \begin{figure}[ht]
  \setlength{\fboxsep}{2.5pt}
  \fbox{
\begin{minipage}{0.46\textwidth}
 {
  \mysize
\begin{tabular}{p{0.95\textwidth}}
\texttt{sumunicodes(A,B) $\leftarrow$ ho_map(A,C,ord),ho_fold(C,B,zero,sum)}\\
\end{tabular}}
\end{minipage}
}
\caption{
\emph{sumunicodes} program which returns the sum of the unicodes of a list of characters.}
\label{fig:sumunicodes}
\end{figure}


\subsection{Q3: Scalability}
Figure \ref{fig:stevie} shows the running times of \name{} on progressively larger programs.
The running time increases exponentially with the size (number of literals) of a program. 
As the size increases, \name{} builds more abstractions, leading to more decision variables in the compress stage.
Note that the running time is the time \name{} needs to find an optimal refactoring and prove optimality.
As Duman\v{c}i\'{c}, Guns, and Cropper \shortcite{knorf} show, for refactoring problems, a solver can quickly find an almost optimal solution but takes a while to find an optimal one.
Overall, these results suggest that the scalability (in terms of proving optimality) of \name{} is limited (\textbf{Q3}).

\subsection{Q4: Transfer Learning}

Experiment 1 explores whether discovering abstractions can improve learning performance on a single domain.
We now explore whether abstractions discovered in one domain can improve performance in different domains.

\paragraph{Domains.}
We use 35 existing tasks which all benefit from higher-order abstractions \cite{metabias,metaho,cretu2022constraint,hopper}.
These tasks are from 7 domains: \emph{chess}, \emph{ascii art}, \emph{string transformations}, \emph{robot strategies}, 
\emph{list manipulation},
\emph{tree manipulation}, and \emph{arithmetic}.
These domains have diverse BK with little overlap.
The appendix contains a description of the domains.

\paragraph{Method.}
Our experimental approach is similar to Experiment 1 but the domains differ in steps 1 and 3. 
In step 1, we use \hopper{} on the tasks from the program synthesis domain. In step 2, we use \name{} to discover abstractions from programs learned in step 1. 
In step 3, we use \hopper{} on tasks in a transfer domain.
%
We infer the type of abstractions discovered by \name{} from the types of the BK in the synthesis domain.
We map the types of abstractions into the transfer domains using a hard-coded type mapping.
We remove abstractions that use a relation which does not exist in the target domain to ensure they can be executed.
The baseline is when we do not apply \name{} in step 2, i.e. \emph{no refactoring}.

\begin{table}[ht]
\centering
\footnotesize
\begin{tabular}{@{}l|cc@{}}
\textbf{Task} &
\textbf{Baseline} & \textbf{\name{}}\\\midrule
\emph{do5times} & 50 $\pm$ 0 & \textbf{100 $\pm$ 0}\\
\emph{line1} & 50 $\pm$ 0 & \textbf{100 $\pm$ 0}\\
\emph{line2} & 50 $\pm$ 0 & \textbf{100 $\pm$ 0}\\
\midrule
\emph{string1} & 50 $\pm$ 0 & \textbf{100 $\pm$ 0}\\
\emph{string2} & 50 $\pm$ 0 & \textbf{100 $\pm$ 0}\\
\emph{string3} & 50 $\pm$ 0 & \textbf{100 $\pm$ 0}\\
\emph{string4} & 50 $\pm$ 0 & \textbf{100 $\pm$ 0}\\
\midrule
\emph{chessmapuntil} & 50 $\pm$ 0 & \textbf{98 $\pm$ 1}\\
\emph{chessmapfilter} & 50 $\pm$ 0 & \textbf{100 $\pm$ 0}\\
\emph{chessmapfilteruntil} & 50 $\pm$ 0 & \textbf{98 $\pm$ 1}\\
\midrule
\emph{droplastk} & 50 $\pm$ 0 & \textbf{100 $\pm$ 0}\\
\emph{encryption} & 50 $\pm$ 0 & \textbf{100 $\pm$ 0}\\
\emph{length} & 80 $\pm$ 12 & \textbf{100 $\pm$ 0}\\
\emph{rotateN} & 50 $\pm$ 0 & \textbf{100 $\pm$ 0}\\
\midrule
\emph{waiter} & 50 $\pm$ 0 & \textbf{100 $\pm$ 0}\\
\end{tabular}
\caption{Predictive accuracies. 
We only include tasks where the two approaches differ.
The full table is in the appendix.
}
\label{tab:transferaccuracies}
\end{table}

\subsubsection{Results}
Table \ref{tab:transferaccuracies} shows the predictive accuracies. The learning times are in the appendix. 
These results show that transferring abstractions (i)  never degrades accuracies, and (ii) can improve accuracies in 5/7 transfer domains. 
A paired t-test confirms ($p<0.01$) the significance of the difference in accuracy for all tasks in Table \ref{tab:transferaccuracies} except \emph{length}.
For instance, \name{} discovers the abstractions \emph{filter} and \emph{map} in the \emph{program synthesis} domain and \hopper{} uses these abstractions for the task \emph{string1} to learn a program which removes lowercase letters and lowercases the remaining letters.
\hopper{} also reuses these abstractions
to learn a solution for the task \emph{chessmapfilter}. Similarly, \hopper{} reuses the abstraction \emph{until} to draw a diagonal line in the \emph{ascii art} domain (Figure \ref{fig:line2}).


\hopper{} struggles on some tasks because \name{} does not discover a helpful abstraction.
For instance, the task \emph{isPalindrome} needs the abstraction \emph{condList}, which returns true if the input list is empty and otherwise calls a predicate on the list.
\name{} does not discover this abstraction because it does not compress the input program.

\hopper{} also struggles on some tasks because of type inconsistencies.
For instance, the task \emph{droplast} involves learning a program which, given a list of lists, drops the last element from each list. \name{} discovers the abstraction \emph{map}. 
However, this abstraction applies to arguments of type \emph{list} whereas \emph{droplast} takes as arguments \emph{lists of lists}.

Overall, these results suggest that higher-order refactoring can improve learning performance in different domains (\textbf{Q4}).

\begin{figure}
   \setlength{\fboxsep}{2.5pt}
  \fbox{
\begin{minipage}{0.45\textwidth}
\mysize
\begin{tabular}{l}
\texttt{line2(A,B) $\leftarrow$ ho_until(A,B,inv\_0,at_right)}\\
\texttt{inv_0(A,B) $\leftarrow$ draw1(A,C),right(C,D),down(D,B)}\\
\end{tabular}
\end{minipage}
}
\caption{\emph{line2} program which draws a diagonal line in an image. 
 The predicate \texttt{inv\_0} is invented by \hopper{}. \label{fig:line2}}
\end{figure}


\section{Conclusions and Limitations}
We introduced an approach that refactors a logic program by discovering higher-order abstractions.
We implemented our approach in \name{}, which formulates this refactoring problem as a COP. 
Our experiments on multiple domains show that higher-order refactoring can drastically improve the performance of an ILP system, namely improving predictive accuracies and reducing learning times.
Our results also show that abstractions discovered in one domain can transfer to different domains. 
For instance, we can discover the abstractions \emph{map}, \emph{filter}, and \emph{fold} in the \emph{program synthesis} domain and use them in the \emph{chess} domain.

\subsubsection{Limitations}

\paragraph{Objective function.} 
Experiment 2 shows that compression alone is not the best metric for identifying abstractions which improve learning performance the most.
Future work should investigate alternative objective functions.

\paragraph{Refactoring time.} Experiment 3 shows that \name{} can optimally refactor programs with around 460 literals in 16 minutes but struggles on larger programs.
Future work should improve scalability, such as improving our COP encoding and using parallel COP solving.
\bibliographystyle{named}
\bibliography{ijcai24}

 \begin{appendices}

\section{Terminology}
\label{sec:bk}
\subsection{Logic Programming}
We assume familiarity with logic programming \cite{lloyd:book} but restate some key relevant notation. A \emph{variable} is a string of characters starting with an uppercase letter. A \emph{predicate} symbol is a string of characters starting with a lowercase letter. The \emph{arity} $n$ of a function or predicate symbol is the number of arguments it takes. A constant symbol is a function or a predicate symbol with arity zero. A variable is \emph{first-order} if it can be bound to a constant symbol or another first-order variable. A variable is \emph{higher-order} if it can be bound to a predicate symbol or another higher-order variable. A term is a variable or a constant symbol. A \emph{first-order atom} is a tuple $p(t_1, ..., t_n)$, where $p$ is a predicate of arity $n$ and $t_1$, ..., $t_n$ are first-order terms. An atom is \emph{ground} if it contains no variables. A \emph{higher-order atom} is a tuple $p(t_1, ..., t_n)$, where $p$ is a predicate of arity $n$ and $t_1$, ..., $t_n$ are terms where at least one $t_i$ is higher-order. A \emph{first-order literal} is a first-order atom or the negation of a first-order atom. A \emph{higher-order literal} is a higher-order atom or the negation of a higher-order atom. A \emph{clause} is a set of literals. The variables in a clause are universally quantified. A clause is higher-order if it contains at least one higher-order literal. A \emph{constraint} is a clause without a positive literal. A \emph{definite} clause is a clause with exactly one positive literal. 
A \emph{program} is a set of definite clauses. 
A program is higher-order if it contains at least one higher-order clause.

The least Herbrand model $M(P)$ of the program $P$ is the set of all ground atomic logical consequences of $P$. 
The least Herbrand model $M(P,B)$ of the programs $P$ and $B$ is $M(P \cup B)$.
In the following, we assume a program $B$ denoting background knowledge and concisely note $M(P,B)$ as $M(P)$.

\section{Correctness}

\label{sec:correctness}
We assume a program $P$ where the definitions do not depend on each other:
\begin{assumption}
\label{bkindependance}
Let $T$ be the head predicate symbols of the program $P$. 
Then every rule in $P$ contains exactly one predicate symbol from $T$.
\end{assumption}

\noindent
Assumption \ref{bkindependance} allows for rules where a predicate symbol may appear multiple times in a rule, i.e. recursive rules.

We show that \name{} generates all abstractions (which are not identical up to renaming of their head predicate symbol) and all instantiations of $P$:
\begin{lemma} \name{} generates all abstractions $\mathcal{A}(P)$ of $P$ which are not identical up to renaming of their head predicate symbol and all instantiations $\mathcal{I}(\mathcal{A}(P))$ of $P$.
\label{lemma:build_all}
\end{lemma}
\begin{proof}

\name{} enumerates every definition $d\in \delta(P)$, of which there are finitely many. 
For each definition $d\in \delta(P)$, \name{} enumerates all subsets of the non-recursive body literals of $d$, of which there are finitely many. Therefore, \name{} builds all abstractions and instantiations. \name{} prunes abstractions that are identical up to renaming of their head predicate symbol.
\end{proof}

\noindent
When we refactor a program, we want to preserve its semantics.
However, we only need to preserve the semantics with respect to head predicate symbols.
Therefore, we reason about the least Herbrand model restricted to a set of predicate symbols:
\begin{definition}[\textbf{Restricted least Herbrand model}]
Let $P$ be a program and $T$ be the head predicate symbols of $P$.
Then the least Herbrand model of $P$ restricted to $T$ is $M_T(P) = \{ a \in M(P) \,|\, \text{the predicate symbol of $a$ is in } T\}$.
\end{definition}

\noindent
We show that a definition has the same \emph{restricted least Herbrand model} than an abstraction and instantiation pair built from it:


\begin{lemma} 
\label{lemma:existence}
Let $d\in \delta(P)$, $h$ be the head predicate symbol of $d$, and $a$ and $i$ be an abstraction and instantiation pair built by \name{} for $d$.
Then $M_h(d) = M_{h}(a \cup i)$.
\end{lemma}
\begin{proof}
We follow Cropper and Tourret \shortcite{reduce} and reason about encapsulated programs. 
The result then follows from the correctness of the unfold operator for first-order logic programs \cite{unfolding,ilp:book}.
\end{proof}

\noindent
We define the \emph{higher-order refactoring} problem:
\begin{definition}[\textbf{Higher-order refactoring problem}]
\label{def:prob_appendix}
Let $P$ be a logic program
and $T$ be the head predicate symbols of $P$.
Then the \emph{higher-order refactoring problem} is to find $Q \subseteq P \cup \mathcal{A}(P) \cup \mathcal{I}(\mathcal{A}(P))$ such that $M_T(Q) = M_T(P)$.
We call $Q$ a \emph{solution} to the refactoring problem.
\end{definition}

\noindent


\noindent
We show that any program output by \name{} is a solution to the refactoring problem:
\begin{proposition}[\textbf{Solution}]
Any program output by \name{} is a solution to the refactoring problem.
\label{prop:solution}
\end{proposition}
\begin{proof}
\name{} outputs a subset of definitions, abstractions, and instantiations of $P$.
Therefore, $Q \subseteq P \cup \mathcal{A}(P) \cup \mathcal{I}(\mathcal{A}(P))$. 

To show $M(Q) = M(P)$, we first show $M(P) \subseteq M(Q)$. 
Let $x \in M(P)$. 
We show $x \in M(Q)$.
If $x \in M(P)$ then $x$ is defined by a definition $d \in \delta(P)$. 
The definition $d$ is either (i) not refactored, or (ii) refactored. 
For case (i), 
since $d$ is not refactored, then $d \subseteq Q$, so $x \in M(Q)$.
For case (ii), since $d$ is refactored, our COP encoding ensures that exactly one instantiation $i$ of $d$ is selected and that the corresponding abstraction $a$ is also selected.
By Lemma \ref{lemma:build_all}, \name{} builds all abstractions and instantiations, so $a$ and $i$ must be built and $a \cup i \subseteq Q$.
By Lemma \ref{lemma:existence}, $M_h(d) = M_h(a \cup i)$, where $h$ is the head predicate symbol of $d$.
Therefore, $x \in M(Q)$.

We now show $M(Q) \subseteq M(P)$. 
Let $x \in M(Q)$. 
We show $x \in M(P)$.
By definition, $x$ can only be defined by (i) a definition, or (ii) an instantiation from $Q$.
For case (i), if $x$ is defined by a definition $d \subseteq Q$, then $d$ must be a non-refactored definition in $P$, which implies that $x \in M(P)$.
For case (ii), assume $x$ is defined by an instantiation $i$. Our COP encoding ensures that exactly one instantiation $i$ of each refactored definition $d \in \delta(P)$ is selected and that the corresponding abstraction $a$ is also selected.
By Lemma \ref{lemma:existence}, $M_h(d) = M_h(a \cup i)$, where $h$ is the head predicate symbol of $d$. 
Therefore, $x \in M(P)$ and $M(Q) \subseteq M(P)$. 
Then $M(Q) = M(P)$, which completes the proof.
\end{proof}

\noindent
Our goal is to perform \emph{optimal refactoring}:
\begin{definition}[\textbf{Optimal refactoring}]\label{def:optprob_appendix}
Let $P$ be a logic program, $T$ be the head predicate symbols of $P$, and $cost$ be a function which maps logic programs to integers.
Then $Q$ is an \emph{optimal} solution when (i) $Q$ is a solution to the refactoring problem, and (ii) there is no $Q' \subseteq  P \cup \mathcal{A}(P) \cup \mathcal{I}(\mathcal{A}(P))$ such that $Q'$ is a solution to the refactoring problem and $cost(Q') < cost(Q)$.
\end{definition}

\noindent
We prove the correctness of \name{}, i.e. \name{} returns an optimal solution to the refactoring problem (Definition \ref{def:optprob_appendix}):
\begin{theorem}[\textbf{Optimal correctness}] \name{} solves the optimal refactoring problem with respect to our objective function.
\end{theorem}
\begin{proof}
By Lemma \ref{lemma:build_all}, \name{} builds all abstractions and instantiations of $P$. Therefore, \name{} can find every model of our COP encoding.
By Proposition \ref{prop:solution}, every model output by \name{} is a solution.
Moreover, the solver finds a solution that minimises our objective function.
\end{proof}

\section{Refactoring}
\label{sec:encoding}
We show an example of higher-order refactoring. 

\subsection{Abstract}
In the \emph{abstract} stage, \name{} builds candidate abstractions. For instance, consider the input program $P$ shown in Figure \ref{fig:input_prog}. This program contains 8 definitions. 
Some abstractions built by \name{} for this program are
shown in Figure \ref{fig:candidateabstractions}.

\begin{figure*}[ht!]
\begin{tabular}{@{}ll@{}}
\multirow{2}{*}{$d_0$}  & 
\texttt{memberzero(A) $\leftarrow$ head(A,B),zero(B)}\\
& \texttt{memberzero(A) $\leftarrow$ tail(A,B),memberzero(B)}\\
\multirow{2}{*}{$d_1$}  & 
\texttt{mapaddone(A,B) $\leftarrow$ empty(A),empty(B)}\\
& \texttt{mapaddone(A,B) $\leftarrow$ head(A,D),tail(A,F),head(B,C),tail(B,E),increment(D,C),mapaddone(F,E)}\\
\multirow{2}{*}{$d_2$}  & 
\texttt{memberodd(A) $\leftarrow$ head(A,B),odd(B)}\\
& \texttt{memberodd(A) $\leftarrow$ tail(A,B),memberodd(B)}\\
\multirow{2}{*}{$d_3$}  & 
\texttt{allnegative(A) $\leftarrow$ empty(A)}\\
&\texttt{allnegative(A) $\leftarrow$ head(A,B),tail(A,C),negative(B),allnegative(C)}\\
\multirow{2}{*}{$d_4$}  & 
\texttt{chartoint(A,B) $\leftarrow$ empty(A),empty(B)}\\
&\texttt{chartoint(A,B) $\leftarrow$ head(A,D),tail(A,F),head(B,C),tail(B,E),ord(D,C),chartoint(F,E)}\\
\multirow{2}{*}{$d_5$}  & 
\texttt{membereven(A) $\leftarrow$ head(A,B),even(B)}\\
& \texttt{membereven(A) $\leftarrow$ tail(A,B),membereven(B)}\\
\multirow{2}{*}{$d_6$}  & 
\texttt{mapcube(A,B) $\leftarrow$ empty(A),empty(B)}\\
& \texttt{mapcube(A,B) $\leftarrow$ head(A,C),tail(A,D),head(B,E),tail(B,F),cube(C,E),mapcube(D,F)}\\
\multirow{2}{*}{$d_7$}  & \texttt{inttobin(A,B) $\leftarrow$ empty(A),empty(B)}\\
& \texttt{inttobin(A,B) $\leftarrow$ head(A,D),tail(A,E),head(B,C),tail(B,F),bin(D,C),inttobin(E,F)}\\
\end{tabular}
\caption{Example input program}
\label{fig:input_prog}
\end{figure*}

\begin{figure*}[ht!]
\begin{tabular}{@{}ll@{}}
\multirow{2}{*}{$a_0$}  & 
\texttt{ho3(A,P) $\leftarrow$ head(A,B),P(B)}\\
&\texttt{ho3(A,P) $\leftarrow$ tail(A,B),ho3(B,P)}\\
\multirow{2}{*}{$a_1$}  & \texttt{ho5(A,P,Q) $\leftarrow$ head(A,B),P(B)}\\
&\texttt{ho5(A,P,Q) $\leftarrow$ Q(A,B),ho5(B,P,Q)}\\
\multirow{2}{*}{$a_2$}  & \texttt{ho6(A,P,Q) $\leftarrow$ P(A,B),Q(B)}\\
&\texttt{ho6(A,P,Q) $\leftarrow$ tail(A,B),ho6(B,P,Q)}\\
\multirow{2}{*}{$a_3$}  & \texttt{ho7(A,P,Q,R) $\leftarrow$ P(A,B),Q(B)}\\
&\texttt{ho7(A,P,Q,R) $\leftarrow$ R(A,B),ho7(B,P,Q,R)}\\
\multirow{2}{*}{$a_4$}  & \texttt{ho8(A,B,P) $\leftarrow$ empty(A),empty(B)}\\
&\texttt{ho8(A,B,P) $\leftarrow$ head(A,C),tail(A,D),head(B,E),tail(B,F),P(C,E),ho8(D,F,P)}\\
\multirow{2}{*}{$a_5$}  & \texttt{ho12(A,B,P,Q) $\leftarrow$ P(A),P(B)}\\
&\texttt{ho12(A,B,P,Q) $\leftarrow$ head(A,C),tail(A,D),head(B,E),tail(B,F),Q(C,E),ho12(D,F,P,Q)}\\
\multirow{2}{*}{$a_6$}  & \texttt{ho13(A,B,P,Q) $\leftarrow$ empty(A),empty(B)}\\
&\texttt{ho13(A,B,P,Q) $\leftarrow$ head(A,C),P(A,D),head(B,E),P(B,F),Q(C,E),ho13(D,F,P,Q)}\\
\multirow{2}{*}{$a_7$}  & \texttt{ho14(A,B,P,Q) $\leftarrow$ empty(A),empty(B)}\\
&\texttt{ho14(A,B,P,Q) $\leftarrow$ P(A,C),tail(A,D),P(B,E),tail(B,F),Q(C,E),ho14(D,F,P,Q)}\\
\multirow{2}{*}{$a_8$}  & \texttt{ho18(A,B,P,Q,R) $\leftarrow$ P(A),P(B)}\\
&\texttt{ho18(A,B,P,Q,R) $\leftarrow$ head(A,C),Q(A,D),head(B,E),Q(B,F),R(C,E),ho18(D,F,P,Q,R)}\\
\multirow{2}{*}{$a_9$}  & \texttt{ho19(A,B,P,Q,R) $\leftarrow$ P(A),P(B)}\\
&\texttt{ho19(A,B,P,Q,R) $\leftarrow$ Q(A,C),tail(A,D),Q(B,E),tail(B,F),R(C,E),ho19(D,F,P,Q,R)}\\
\multirow{2}{*}{$a_{10}$}  & \texttt{ho20(A,B,P,Q,R) $\leftarrow$ empty(A),empty(B)}\\
&\texttt{ho20(A,B,P,Q,R) $\leftarrow$ P(A,C),Q(A,D),P(B,E),Q(B,F),R(C,E),ho20(D,F,P,Q,R)}\\
\end{tabular}
\caption{Example candidate abstractions built by \name{} given the input program shown in Figure \ref{fig:input_prog}.}
\label{fig:candidateabstractions}
\end{figure*}

\subsection{Compress}
Figure \ref{fig:output} shows the program output by \name{} given the input program shown in Figure \ref{fig:input_prog}. \name{} has selected the candidate abstractions \emph{$a_0$} and \emph{$a_4$}. No abstraction has been selected for the definition \emph{$d_3$}. The output program has size 37 while the input program has size 65.

\begin{figure*}[ht!]
\begin{tabular}{@{}l@{}}
\texttt{ho3(A,P) $\leftarrow$ head(A,B),P(B)}\\
\texttt{ho3(A,P) $\leftarrow$ tail(A,B),ho3(B,P)}\\
\texttt{memberzero(A) $\leftarrow$ ho3(A,zero)}\\
\texttt{memberodd(A) $\leftarrow$ ho3(A,odd)}\\
\texttt{membereven(A) $\leftarrow$ ho3(A,even)}\\
\texttt{ho8(A,B,P) $\leftarrow$ empty(A),empty(B)}\\
\texttt{ho8(A,B,P) $\leftarrow$ head(A,C),tail(A,D),head(B,E),tail(B,F),P(C,E),ho8(D,F,P)}\\
\texttt{mapaddone(A,B) $\leftarrow$ ho8(A,B,increment)}\\
\texttt{chartoint(A,B) $\leftarrow$ ho8(A,B,ord)}\\
\texttt{mapcube(A,B) $\leftarrow$ ho8(A,B,cube)}\\
\texttt{inttobin(A,B) $\leftarrow$ ho8(A,B,bin)}\\
\texttt{allnegative(A) $\leftarrow$ empty(A)}\\
\texttt{allnegative(A) $\leftarrow$ head(A,B),tail(A,C),negative(B),allnegative(C)}\\
\end{tabular}
\caption{Refactored program output by \name{} given the input program from Figure \ref{fig:input_prog} and the candidate abstractions from Figure \ref{fig:candidateabstractions}.}
\label{fig:output}
\end{figure*}

\section{Experiments}
\label{sec:exp_domains}

\subsection{Experimental domains}

\paragraph{Program Synthesis.} This dataset includes list transformation tasks. It involves learning recursive programs which has been identified as a difficult challenge for ILP systems \cite{ilp20}. We design tasks to purposely require higher-order constructs. Table \ref{fig:programsynthesistasks} shows example first-order and higher-order solutions for some of the tasks.

\paragraph{Chess tactics.} The task is to induce chess strategies, such as maintaining a wall of pawns to support promotion \cite{metaho}. Examples are pairs of input-output states.

\paragraph{Ascii art.} The goal is to induce programs that manipulate images \cite{cretu2022constraint}.
Figure \ref{fig:ascii_exs} shows a training example for the problem of drawing a diagonal line (\emph{line2}).
Figure \ref{fig:ascii} shows an example of target hypothesis.

\paragraph{String transformations.} This dataset includes real-world string transformation. We constrain systems to learn functional programs \cite{metabias} to compensate for the lack of negative examples. 

\paragraph{Robot strategies.} The task is to learn a strategy for a robot to pour tea and coffee at a dinner table \cite{metaho}. Examples are pairs of initial and final state. In the initial state, the cups are empty and each guest has a preference for tea or coffee. In the final state, the cups are filled with the preferred drink. Figure \ref{fig:waiter} shows an example of hypothesis.

\paragraph{List manipulation.} We use 7 tasks introduced by Cropper and Morel \shortcite{popper}, 2 tasks introduced by Cropper et al. \shortcite{metaho} and 8 tasks introduced by Purgal et al. \shortcite{hopper}. 

\paragraph{Tree manipulation.} We use 3 tasks introduced by Purgal et al. \shortcite{hopper}: finding the depth of a tree (\emph{depth}, checking whether a given list is a branch of the tree (\emph{isBranch}) and check whether a tree is a sub-tree of the first argument (\emph{isSubTree}).

\paragraph{Arithmetic.} We use 2 tasks introduced by Purgal et al. \shortcite{hopper}: adding a number $n$ to every element of a list with no addition predicate in
the BK (\emph{addN}) and multiplying two numbers with no addition predicate in the
BK (\emph{multFromSucc}).

\def\pixelsxa{
  {0,0,1,0},
  {0,0,0,1},
  {0,1,0,0},
  {0,1,0,0}%
}
\def\pixelsya{
  {1,0,1,0},
  {0,1,0,1},
  {0,1,1,0},
  {0,1,0,1}%
}

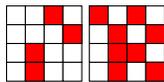
\begin{figure}[ht]
\centering
\centering
\begin{tikzpicture}[scale=0.25]
  \foreach \line [count=\y] in \pixelsxa {
    \foreach \pix [count=\x] in \line {
      \draw[fill=pixel \pix] (\x,-\y) rectangle +(1,1);
        \ifthenelse{\pix = 0}
        {}
        {\draw[fill=red] (\x,-\y) rectangle +(1,1);}

    }
  }
\end{tikzpicture}
\begin{tikzpicture}[scale=0.25]
  \foreach \line [count=\y] in \pixelsya {
    \foreach \pix [count=\x] in \line {
      \draw[fill=pixel \pix] (\x,-\y) rectangle +(1,1);
        \ifthenelse{\pix = 0}
        {}
        {\draw[fill=red] (\x,-\y) rectangle +(1,1);}

    }
  }
\end{tikzpicture}
\caption{Ascii art example input-output pair.}
\label{fig:ascii_exs}
\end{figure}

\begin{figure}[ht]
\centering
\subfloat[Ascii art hypothesis without \name{}]{
\begin{tabular}{l}
\texttt{line2(A,B)$\leftarrow$ draw1(A,C),move$\_$right(C,D),}\\
\hspace{66pt}\texttt{move$\_$down(D,B),at$\_$end(B)}\\
\texttt{line2(A,B)$\leftarrow$ draw1(A,C),move$\_$right(C,D),}\\
\hspace{66pt}\texttt{move$\_$down(D,E),line2(E,B)}\\

\end{tabular}}\\

\subfloat[Ascii art with \name{}. The symbol \tw{inv} is invented by \hopper{}.]{
\begin{tabular}{l}
\texttt{line2(A,B) $\leftarrow$ until(A,B,inv,at$\_$right).
}\\
\texttt{inv(A,B) $\leftarrow$ draw1(A,C),move$\_$right(C,D),}\\
\hspace{71pt}\texttt{move$\_$down(D,E)}\\
\end{tabular}}\\
\caption{
Example of ascii art hypothesis.
}
\label{fig:ascii}
\end{figure}

\begin{figure}[ht]
\centering
\subfloat[Waiter hypothesis without \name{}]{
\begin{tabular}{l}
\texttt{f(A,B)$\leftarrow$ f1(A,B),at$\_$end(B)}\\
\texttt{f(A,B)$\leftarrow$ wants$\_$tea(A),pour$\_$tea(A,C))}\\
\hspace{44pt}\texttt{move$\_$right(C,B),f(D,B)}\\
\texttt{f(A,B)$\leftarrow$ wants$\_$coffee(A),pour$\_$coffee(A,C))}\\
\hspace{44pt}\texttt{move$\_$right(C,B),f(D,B)}\\
\end{tabular}}\\

\subfloat[Waiter hypothesis with \name{}. The symbol \tw{inv} is invented by \hopper{}.]{
\begin{tabular}{l}
\texttt{f(A,B) $\leftarrow$ until(A,B,at$\_$end,inv)}\\
\texttt{inv(A,B) $\leftarrow$ wants$\_$tea(A),pour$\_$tea(A,C)}\\
\hspace{71pt}\texttt{move$\_$right(C,B)}\\
\texttt{inv(A,B) $\leftarrow$ wants$\_$coffee(A),pour$\_$coffee(A,C),}\\
\hspace{71pt}\texttt{move$\_$right(C,B)}\\
\end{tabular}}\\
\caption{
Example of robot strategy hypothesis.
}
\label{fig:waiter}
\end{figure}

\subsection{Systems}
We use a version of \hopper{} based on \popper{} 2.0.0\footnote{\url{https://github.com/logic-and-learning-lab/Popper/releases/tag/v2.0.0}}.

\subsection{Abstractions}
\label{sec:usualabstractions}
Table \ref{tab:abstractionsswiprolog} shows the abstractions used in our experiments as a baseline. These abstractions are from the SWI-Prolog library \emph{apply}\footnote{\url{https://www.swi-prolog.org/pldoc/man?section=apply}}.

\begin{table*}[ht!]
\footnotesize
\centering
\begin{tabular}{@{}lll@{}}
\textbf{Name} & \textbf{Abstraction} & \textbf{Bias}\\
\midrule
\multirow{5}{*}{\emph{include}} &\texttt{include([], $\_$, []).} & \texttt{body$\_$pred(include,3,ho).}
\\
&\texttt{include([H|T], Included, P) :-} & \texttt{type(include,(list,list,(element,))).}\\
&\hspace{10pt}\texttt{(call(P, H) -> Included=[H|Included1];} & \texttt{direction(exclude,(in,out,(in,))).}\\
&\hspace{10pt}\texttt{Included = Included1),} & \\
&\hspace{10pt}\texttt{include(T, Included1, P).} & \\\midrule

\multirow{5}{*}{\emph{exclude}} &\texttt{exclude([], $\_$, []).} & \texttt{body$\_$pred(exclude,3,ho).}
\\
&\texttt{exclude([H|T], Included, P) :-} & \texttt{type(exclude,(list,list,(element,))).}\\
&\hspace{10pt}\texttt{(call(P, H) -> Included = Included1;} & \texttt{direction(exclude,(in,out,(in,))).}\\
&\hspace{10pt}\texttt{Included=[H|Included1]),} &\\
&\hspace{10pt}\texttt{exclude(T, Included1, P).} & \\\midrule

\multirow{4}{*}{\emph{maplist}} &
\texttt{maplist([], [], $\_$).} & \texttt{body$\_$pred(maplist,3,ho).}\\
&\texttt{maplist([H1|T1], [H2|T2], Goal) :-} & \texttt{type(maplist,(list,list,(element,element))).}\\
&\hspace{10pt}\texttt{
call(Goal, H1, H2),} & \texttt{direction(maplist,(in,out,(in,out))).}\\
&\hspace{10pt}\texttt{
maplist(T1, T2, Goal).} & \\
\midrule

\multirow{5}{*}{\emph{convlist}} &
\texttt{convlist([], [], $\_$).} & \texttt{body$\_$pred(convlist,3,ho).}\\
&\texttt{convlist([H0|T0], ListOut, Goal) :-} & \texttt{type(convlist,(list,list,(element,element))).}\\
&\hspace{10pt}\texttt{
call(Goal, H0, H)->  ListOut = [H|T],} & \texttt{direction(convlist,(in,out,(in,out))).}\\
&\hspace{10pt}\texttt{convlist(T0, T, Goal);} & \\
&\hspace{10pt}\texttt{convlist(T0, ListOut, Goal).} & \\
\midrule

\multirow{4}{*}{\emph{foldl}} &
\texttt{foldl([], V, V, $\_$).} & \texttt{body$\_$pred(foldl,4,ho).}\\
&\texttt{foldl([H|T], V0, V, Goal) :-} & \texttt{type(foldl,(list,element,element,(element,element,element))).}\\
&\hspace{10pt}\texttt{
call(Goal, H, V0, V1),} & \texttt{direction(foldl,(in,in,out,(in,in,out))).}\\
&\hspace{10pt}\texttt{
foldl(T, V1, V, Goal).} & \\
\midrule

\multirow{5}{*}{\emph{partition}} &
\texttt{partition([], [], [], $\_$).} & \texttt{body$\_$pred(partition,4,ho).}\\
&\texttt{partition([H|T], Pred, Incl, Excl) :-} & \texttt{type(partition,(list,list,list,(element,))).}\\
&\hspace{10pt}\texttt{(call(Pred, H) ->  Incl = [H|I], } & \texttt{direction(partition,(in,out,out,(in,))).}\\
&\hspace{10pt}\texttt{partition(T, Pred, I, Excl);} & \\
&\hspace{10pt}\texttt{Excl = [H|E],
   partition(T, Pred, Incl, E)).} & \\
\midrule
\multirow{4}{*}{\emph{scanl}} &
\texttt{scanl([], $\_$, [], $\_$).} & \texttt{body$\_$pred(scanl,4,ho).}\\
&\texttt{scanl([H|T], V, [VH|VT], Goal) :-} & \texttt{type(scanl,(list,element,list,(element,element,element))).}\\
&\hspace{10pt}\texttt{
call(Goal, H, V, VH),} & \texttt{direction(scanl,(in,in,out,(in,in,out))).}\\
&\hspace{10pt}\texttt{
scanl(T, VH, VT, Goal).} & \\

\end{tabular}
\caption{Usual abstractions from the SWI-Prolog library \emph{apply} used in our experiments.}
\label{tab:abstractionsswiprolog}
\end{table*}

\section{Experimental results}
\label{sec:outputs}
Figure \ref{fig:abstractions} shows examples of abstractions discovered by \name{} on the program synthesis domain. Table \ref{fig:programsynthesistasks} shows some examples of programs learned by \name{} with and without refactoring. Figure \ref{fig:programs} shows some examples of programs learned by \hopper{} when using abstractions discovered by \name{}.

Tables \ref{tab:transferaccuracies_appendix} and \ref{tab:transfertimes} show the predictive accuracies and the learning times for our transfer learning experiment. 

\begin{figure*}[ht!]
\begin{lstlisting}
%% repeat 5
ho_1(A,B,P) :- P(A,C),P(C,D),P(D,E),P(E,F),P(F,B)

%% do until
ho_6(A,B,P,Q) :- P(A,B),head(B,C),Q(C)
ho_6(A,B,P,Q) :- P(A,C),ho_6(C,B,P,Q)

%% progressive list
ho_16(A,P) :- tail(A,B),empty(B)
ho_16(A,P) :- head(A,B),tail(A,C),ho_16(C,P),head(C,D),P(B,D)

% member
ho_31(A,P) :- head(A,B),P(B)
ho_31(A,P) :- tail(A,B),ho_31(B,P)

%% fold
ho_46(A,B,P,Q) :- empty(A),P(B)
ho_46(A,B,P,Q) :- head(A,C),tail(A,D),ho_46(D,E,P,Q),Q(C,E,B)

% all
ho_73(A,P) :- empty(A)
ho_73(A,P) :- head(A,B),tail(A,C),P(B),ho_73(C,P)

%% map
ho_89(A,B,P) :- empty(A),empty(B)
ho_89(A,B,P) :- head(A,C),tail(A,D),P(C,E),ho_89(D,F,P),head(B,E),tail(B,F)

%% count
ho_185(A,B,P,Q) :- empty(A),zero_int(B)
ho_185(A,B,P,Q) :- head(A,C),tail(A,D),P(C),ho_185(D,B,P,Q)
ho_185(A,B,P,Q) :- head(A,C),tail(A,D),Q(C),ho_185(D,E,P,Q),my_increment(E,B)

%% filter
ho_517(A,B,P,Q) :- empty(A),empty(B)
ho_517(A,B,P,Q) :- head(A,C),tail(A,D),P(C),ho_517(D,B,P,Q)
ho_517(A,B,P,Q) :- tail(A,C),ho_517(C,D,P,Q),tail(B,D),head(B,E),Q(E)

%% iterate
ho_542(A,B,C,P) :- zero(A),eq(B,C)
ho_542(A,B,C,P) :- decrement(A,D),ho_542(D,B,E,P),P(E,C)
\end{lstlisting}
\caption{Example of abstractions discovered by \name{} on the \emph{program synthesis} domain.}
\label{fig:abstractions}
\end{figure*}

\begin{table*}[ht!]
\footnotesize
\centering
\begin{tabular}{@{}lll@{}}
\textbf{Task} & \textbf{First-order hypothesis} & \textbf{Higher-order hypothesis}\\
\midrule
\multirow{3}{*}{\emph{allzero}} &
\texttt{allzero(A)$\leftarrow$ empty(A).} & \texttt{allzero(A)$\leftarrow$ ho\_73(A,zero).}\\
& \texttt{allzero(A)$\leftarrow$ head(A,B),tail(A,C),}\\
& \hspace{56pt} \texttt{zero(B),allzero(C).} & \\
\midrule

\multirow{2}{*}{\emph{member2}} & \texttt{member2(A)$\leftarrow$ head(A,B),decrement(B,C),one(C).} & 
\texttt{member2(A)$\leftarrow$ ho\_31(A,inv\_1).}\\
& \texttt{member2(A)$\leftarrow$ tail(A,B),member2(B).} & \texttt{inv\_1(A)$\leftarrow$ decrement(A,B),one(B).}\\ 
\midrule

\multirow{2}{*}{\emph{dropfirst5}} & \texttt{dropfirst5(A,B)$\leftarrow$ tail(A,E),tail(E,F),} & \texttt{dropfirst5(A,B)$\leftarrow$ ho\_1(A,B,tail).}\\
& \hspace{79pt} \texttt{tail(F,D),tail(D,C),tail(C,B).} & \\
\midrule

\multirow{3}{*}{\emph{counteven}} &
\texttt{counteven(A,B) $\leftarrow$ empty(A),zero(B).} & \texttt{counteven(A,B)$\leftarrow$ ho\_185(A,B,even,odd).}\\
& \texttt{counteven(A,B)$\leftarrow$ head(A,C),odd(C),} &\\
& \hspace{74pt} \texttt{tail(A,D),counteven(D,B).} &\\
& \texttt{counteven(A,B)$\leftarrow$ head(A,D),even(D),tail(A,C),} &\\
& \hspace{74pt} \texttt{counteven(C,E),increment(E,B).} & \\\midrule

\multirow{3}{*}{\emph{sumlist}} & 
\texttt{sumlist(A,B)$\leftarrow$ empty(A),zero(B).} &
\texttt{sumlist(A,B)$\leftarrow$ ho\_46(A,B,zero,sum).}\\
& \texttt{sumlist(A,B)$\leftarrow$ head(A,C),tail(A,E),sumlist(E,D),} & \\
& \hspace{67pt}\texttt{sum(C,D,B).} & \\\midrule

\multirow{3}{*}{\emph{sorteddecr}} & \texttt{sorteddecr(A)$\leftarrow$ tail(A,B),empty(B).} & \texttt{sorteddecr(A)$\leftarrow$ ho\_16(A,geq).}\\
& \texttt{sorteddecr(A)$\leftarrow$ head(A,B),tail(A,C),sorteddecr(C),} & \\
&  \hspace{70pt} \texttt{head(C,D),geq(B,D).} & \\ \midrule

\multirow{3}{*}{\emph{inttobin}} & \texttt{inttobin(A,B)$\leftarrow$ empty(A),empty(B).} & \texttt{inttobin(A,B)$\leftarrow$ ho\_89(A,B,bin).}\\
& \texttt{inttobin(A,B)$\leftarrow$ head(A,C),tail(A,F),head(B,D),} & \\
&  \hspace{70pt} \texttt{tail(B,E),bin(C,D),inttobin(F,E).} & \\ \midrule

\multirow{3}{*}{\emph{filtereven}} & \texttt{filtereven(A,B)$\leftarrow$ empty(A),empty(B).} & \texttt{filtereven(A,B)$\leftarrow$ ho\_517(A,B,even,odd).}\\
& \texttt{filtereven(A,B)$\leftarrow$ cons3(A,C,D),even(C),filtereven(D,B).} & \\
& \texttt{filtereven(A,B)$\leftarrow$ cons3(A,C,D),odd(C),} & \\
& \hspace{79pt} \texttt{filtereven(D,E),cons(C,E,B).} & \\ \midrule

\multirow{2}{*}{\emph{iteratedropk}} & \texttt{iteratedropk(A,B,C)$\leftarrow$ zero(A),eq(B,C).}
& \texttt{iteratedropk(A,B,C)$\leftarrow$ ho\_542(A,B,C,tail).} \\ 
& \texttt{iteratedropk(A,B,C)$\leftarrow$ decrement(A,E),tail(B,D),} & \\
& \hspace{97pt} \texttt{iteratedropk(E,D,C).} & \\
\end{tabular}
\caption{Example first-order and higher-order target hypotheses for some of the tasks in the \emph{program synthesis} domain.}
\label{fig:programsynthesistasks}
\end{table*}

\begin{figure*}[ht]
\centering
\begin{tabular}{l}
\texttt{mapaddthree(A,B)$\leftarrow$ ho\_89(A,B,inv\_0).}\\
\texttt{inv\_0(A,B)$\leftarrow$ increment(A,C),increment(C,D),increment(D,B).}\\\midrule

\texttt{encrypt(A,B)$\leftarrow$ ho\_89(A,B,inv\_0).}\\
\texttt{inv\_0(A,B)$\leftarrow$ ord(A,C),increment(C,D),bin(D,B).}\\\midrule

\texttt{filternegativemaptriple(A,B)$\leftarrow$  ho\_517(A,C,negative,positive),ho\_89(C,B,triple).}\\\midrule

\texttt{member0and2(A)$\leftarrow$  ho\_31(A,inv\_0),ho\_31(A,zero).}\\
\texttt{inv\_0(A)$\leftarrow$ decrement(A,B),one(B).}\\\midrule

\texttt{mapaddk(A,B,C)$\leftarrow$ ho\_542(A,B,C,inv\_1)} \\
\texttt{inv\_1(A,B)$\leftarrow$ ho\_89(A,B,increment)} \\\midrule

\texttt{allpositiveallodd(A)$\leftarrow$ ho\_73(A,odd),ho\_73(A,positive).}\\\midrule

\texttt{seqstep3decr(A):- ho\_16(A,inv\_0).}\\
\texttt{inv\_0(A,B):- decrement(A,C),decrement(C,D),decrement(D,B).}\\\midrule

\texttt{iteratedrop4k(A,B,C):- ho\_5(A,B,C,inv\_0).}\\
\texttt{inv\_0(A,B):- tail(A,C),tail(C,D),tail(D,E),tail(E,B).}
\end{tabular}\\
\caption{
Examples of hypotheses learned by \hopper{} when using abstractions discovered by \name{}.
}
\label{fig:programs}
\end{figure*}

\begin{table}[ht]
\centering
\small
\begin{tabular}{l|cc}
\textbf{Task} &
\textbf{Baseline} & \textbf{\name{}}\\\midrule
\emph{chessmapuntil} & 50 $\pm$ 0 & \textbf{98 $\pm$ 1}\\
\emph{chessmapfilter} & 50 $\pm$ 0 & \textbf{100 $\pm$ 0}\\
\emph{chessmapfilteruntil} & 50 $\pm$ 0 & \textbf{98 $\pm$ 1}\\
\emph{chess} & \textbf{50 $\pm$ 0} & \textbf{50 $\pm$ 0}\\
\midrule
\emph{do5times} & 50 $\pm$ 0 & \textbf{100 $\pm$ 0}\\
\emph{line1} & 50 $\pm$ 0 & \textbf{100 $\pm$ 0}\\
\emph{line2} & 50 $\pm$ 0 & \textbf{100 $\pm$ 0}\\
\midrule
\emph{string1} & 50 $\pm$ 0 & \textbf{100 $\pm$ 0}\\
\emph{string2} & 50 $\pm$ 0 & \textbf{100 $\pm$ 0}\\
\emph{string3} & 50 $\pm$ 0 & \textbf{100 $\pm$ 0}\\
\emph{string4} & 50 $\pm$ 0 & \textbf{100 $\pm$ 0}\\
\midrule
\emph{waiter} & 50 $\pm$ 0 & \textbf{100 $\pm$ 0}\\
\emph{waiter2} & \textbf{50 $\pm$ 0} & \textbf{50 $\pm$ 0}\\
\midrule
\emph{alleven} & \textbf{100 $\pm$ 0} & \textbf{100 $\pm$ 0}\\
\emph{allseqN} & \textbf{50 $\pm$ 0} & \textbf{50 $\pm$ 0}\\
\emph{droplast} & \textbf{50 $\pm$ 0} & \textbf{50 $\pm$ 0}\\
\emph{droplastk} & 50 $\pm$ 0 & \textbf{100 $\pm$ 0}\\
\emph{dropk} & \textbf{100 $\pm$ 0} & \textbf{100 $\pm$ 0}\\
\emph{encryption} & 50 $\pm$ 0 & \textbf{100 $\pm$ 0}\\
\emph{finddup} & \textbf{50 $\pm$ 0} & \textbf{50 $\pm$ 0}\\
\emph{firstHalf} & \textbf{50 $\pm$ 0} & \textbf{50 $\pm$ 0}\\
\emph{isPalindrome} & \textbf{50 $\pm$ 0} & \textbf{50 $\pm$ 0}\\
\emph{lastHalf} & \textbf{50 $\pm$ 0} & \textbf{50 $\pm$ 0}\\
\emph{length} & 80 $\pm$ 12 & \textbf{100 $\pm$ 0}\\
\emph{member} & \textbf{100 $\pm$ 0} & \textbf{100 $\pm$ 0}\\
\emph{of1And2} & \textbf{50 $\pm$ 0} & \textbf{50 $\pm$ 0}\\
\emph{repeatN} & \textbf{50 $\pm$ 0} & \textbf{50 $\pm$ 0}\\
\emph{reverse} & \textbf{50 $\pm$ 0} & \textbf{50 $\pm$ 0}\\
\emph{rotateN} & 50 $\pm$ 0 & \textbf{100 $\pm$ 0}\\
\emph{sorted} & \textbf{100 $\pm$ 0} & \textbf{100 $\pm$ 0}\\\midrule
\emph{depth} & \textbf{50 $\pm$ 0} & \textbf{50 $\pm$ 0}\\
\emph{isBranch} & \textbf{100 $\pm$ 0} & \textbf{100 $\pm$ 0}\\
\emph{isSubTree} & \textbf{50 $\pm$ 0} & \textbf{50 $\pm$ 0}\\
\midrule
\emph{addN} & \textbf{50 $\pm$ 0} & \textbf{50 $\pm$ 0}\\
\emph{multFromSuc} & \textbf{50 $\pm$ 0} & \textbf{50 $\pm$ 0}\\
\end{tabular}
\caption{Predictive accuracies. 
}
\label{tab:transferaccuracies_appendix}
\end{table}

\begin{table}[ht]
\centering
\small
\begin{tabular}{l|cc}
\textbf{Task}
&
\textbf{Baseline} & \textbf{ \name{}}\\\midrule
\emph{chessmapuntil} & \textbf{0 $\pm$ 0} & 4 $\pm$ 0\\
\emph{chessmapfilter} & \textbf{2 $\pm$ 0} & 4 $\pm$ 1\\
\emph{chessmapfilteruntil} & \textbf{1 $\pm$ 0} & 3 $\pm$ 0\\
\emph{chess} & \textbf{0 $\pm$ 0} & \emph{timeout}\\\midrule
\emph{do5times} & \emph{timeout} & \textbf{408 $\pm$ 91}\\
\emph{line1} & \emph{timeout} & \textbf{305 $\pm$ 13}\\
\emph{line2} & \emph{timeout} & \textbf{339 $\pm$ 29}\\
\midrule
\emph{string1} & \emph{timeout} & \textbf{267 $\pm$ 161}\\
\emph{string2} & \emph{timeout} & \textbf{6 $\pm$ 1}\\
\emph{string3} & \emph{timeout} & \textbf{708 $\pm$ 33}\\
\emph{string4} & \emph{timeout} & \textbf{91 $\pm$ 36}\\
\midrule
\emph{waiter} & \emph{timeout} & \textbf{5 $\pm$ 1}\\
\emph{waiter2} & \emph{timeout} & \emph{timeout}\\
\midrule
\emph{alleven} & \textbf{1 $\pm$ 0} & \textbf{1 $\pm$ 0}\\
\emph{allseqN} & \emph{timeout} & \emph{timeout}\\
\emph{droplast} & \emph{timeout} & \emph{timeout}\\
\emph{droplastk} & \emph{timeout} & \textbf{8 $\pm$ 2}\\
\emph{dropk} & 19 $\pm$ 6 & \textbf{1 $\pm$ 0}\\
\emph{encryption} & \emph{timeout} & \textbf{175 $\pm$ 42}\\
\emph{finddup} & \emph{timeout} & \emph{timeout}\\
\emph{firstHalf} & \emph{timeout} & \emph{timeout}\\
\emph{isPalindrome} & \emph{timeout} & \emph{timeout}\\
\emph{lastHalf} & \emph{timeout} & \emph{timeout}\\
\emph{length} & \textbf{5 $\pm$ 1} & 10 $\pm$ 1\\
\emph{member} & \textbf{0 $\pm$ 0} & 1 $\pm$ 0\\
\emph{of1And2} & \emph{timeout} & \emph{timeout}\\
\emph{repeatN} & \emph{timeout} & \emph{timeout}\\
\emph{reverse} & \emph{timeout} & \emph{timeout}\\
\emph{rotateN} & \emph{timeout} & \textbf{56 $\pm$ 15}\\
\emph{sorted} & 14 $\pm$ 1 & \textbf{11 $\pm$ 1}\\\midrule
\emph{depth} & \emph{timeout} & \emph{timeout}\\
\emph{isBranch} & \textbf{0 $\pm$ 0} & 4 $\pm$ 0\\
\emph{isSubTree} & \emph{timeout} & \emph{timeout}\\
\midrule
\emph{addN} & \emph{timeout} & \emph{timeout}\\
\emph{multFromSuc} & \emph{timeout} & \emph{timeout}\\
\end{tabular}
\caption{Learning times. The error is standard error.
}
\label{tab:transfertimes}
\end{table}

\end{appendices}

\end{document}